\documentclass[pdflatex,iicol,sn-mathphys-ay]{sn-jnl}

\usepackage{graphicx}%
\usepackage{multirow}%
\usepackage{amsmath,amssymb,amsfonts}%
\usepackage{amsthm}%
\usepackage{mathrsfs}%
\usepackage[title]{appendix}%
\usepackage{xcolor}%
\usepackage{textcomp}%
\usepackage{manyfoot}%
\usepackage{booktabs}%
\usepackage{algorithm}%
\usepackage{algorithmicx}%
\usepackage{algpseudocode}%
\usepackage{listings}%
\usepackage{caption}
\usepackage{subcaption}

\theoremstyle{thmstyleone}%
\theoremstyle{thmstyletwo}%

\theoremstyle{thmstylethree}%

\begin{document}

\title[Article Title]{Towards PerSense\texttt{++}: Advancing Training-Free Personalized Instance Segmentation in Dense Images}


\author{\fnm{Muhammad Ibraheem} \sur{Siddiqui}}

\author{\fnm{Muhammad Umer} \sur{Sheikh}}

\author{\fnm{Hassan} \sur{Abid}}

\author{\fnm{Kevin} \sur{Henry}}

\author{\fnm{Muhammad Haris} \sur{Khan}}

\affil[]{\orgdiv{Department of Computer Vision}, \orgname{Mohamed Bin Zayed University of Artificial Intelligence}, \orgaddress{\city{Abu Dhabi}, \country{UAE}}}




\abstract{Segmentation in dense visual scenes poses significant challenges due to occlusions, background clutter, and scale variations. To address this, we introduce PerSense, an end-to-end, training-free, and model-agnostic one-shot framework for \textbf{Per}sonalized instance \textbf{S}egmentation in d\textbf{ense} images. PerSense employs a novel Instance Detection Module (IDM) that leverages density maps (DMs) to generate instance-level candidate point prompts, followed by a Point Prompt Selection Module (PPSM) that filters false positives via adaptive thresholding and spatial gating. A feedback mechanism further enhances segmentation by automatically selecting effective exemplars to improve DM quality. We additionally present PerSense\texttt{++}, an enhanced variant that incorporates three additional components to improve robustness in cluttered scenes: (i) a diversity-aware exemplar selection strategy that leverages feature and scale diversity for better DM generation; (ii) a hybrid IDM combining contour and peak-based prompt generation for improved instance separation within complex density patterns; and (iii) an Irrelevant Mask Rejection Module (IMRM) that discards spatially inconsistent masks using outlier analysis. Finally, to support this underexplored task, we introduce PerSense-D, a dedicated benchmark for personalized segmentation in dense images. Extensive experiments across multiple benchmarks demonstrate that PerSense\texttt{++} outperforms existing methods in dense settings. Code is available at \href{https://github.com/Muhammad-Ibraheem-Siddiqui/PerSense}{GitHub}.}




\maketitle

\begin{figure*}[t]
    \centering
    \includegraphics[width=0.97\linewidth]{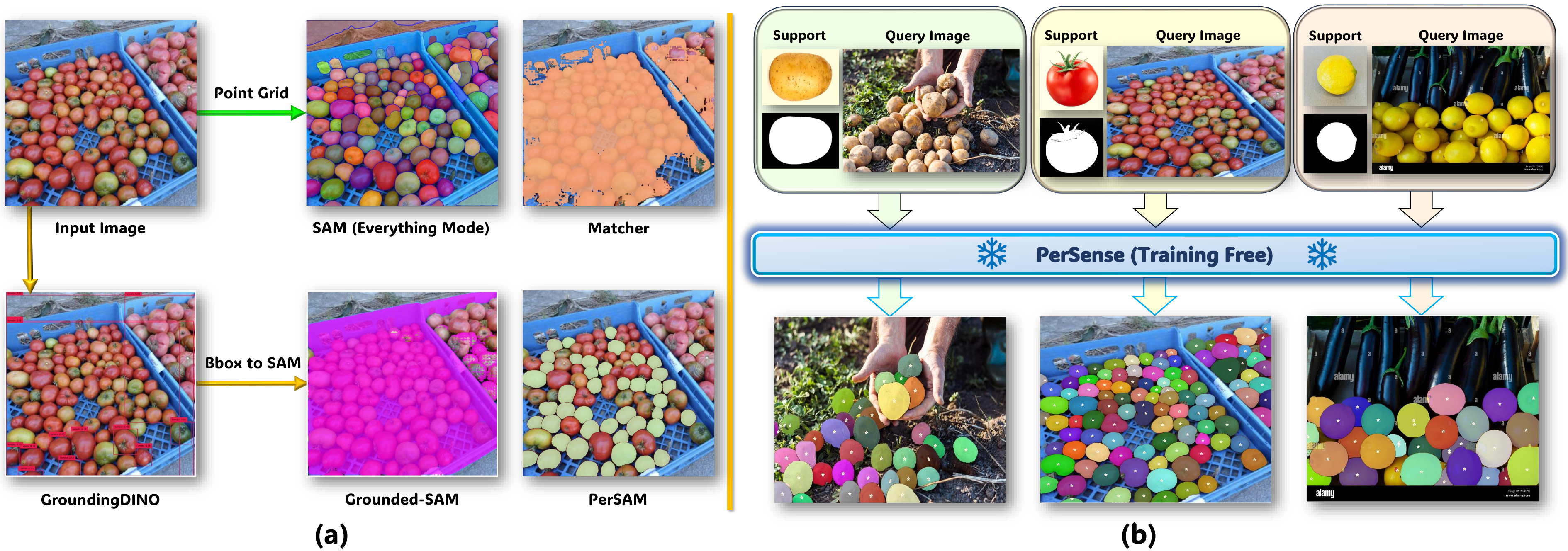}
    \caption{(a) Recent methods struggle in dense scenes, while SAM's “everything mode” segments both foreground and background, indiscriminately. (b) Introducing PerSense: a \textbf{training-free, model-agnostic one-shot framework} for personalized instance segmentation in dense images.}
    \label{fig:introfig}
\end{figure*}

\section{Introduction}\label{intro}


Imagine an industrial setting where the goal is to automate quality control for vegetables, such as potatoes, using vision sensors. We refer to this task as \textit{personalized instance segmentation in dense images}, extending the concept of personalized segmentation introduced in~\cite{zhang2023personalize}. The term \textit{personalized} refers to the segmentation of a specific visual category / concept within an image. Our task setting focuses on personalized instance segmentation, particularly in \textit{dense scenarios}. A natural way to address this problem is leveraging state-of-the-art (SOTA) foundation models. Segment Anything Model (SAM)~\citep{kirillov2023segment} introduced a prompt-driven segmentation framework but struggles to capture distinct visual concepts~\citep{zhang2023personalize}. Its “everything mode” segments all objects via a point grid, including irrelevant background (Fig.\ref{fig:introfig}). While manual prompts can isolate instances, the process is labor-intensive. Automation via box prompts has been explored in Grounded-SAM~\citep{ren2024grounded}, which uses GroundingDINO~\citep{liu2023grounding} detections as input to SAM. However, bounding boxes are limited by box shape, occlusions, and orientation of objects~\citep{zand2021oriented}. A standard axis-aligned box for an object may include portions of adjacent instances. Additionally, non-max suppression (NMS) may merge closely positioned instances. While bipartite matching in DETR~\citep{carion2020end} alleviates this, box-based detections remain hindered by scale variations, occlusions, and clutter, challenges that intensify in dense scenes~\citep{wan2019adaptive} (Fig.~\ref{fig:introfig}).

Point-based prompting, mostly based on manual user input, is generally better than box-based prompting for tasks that require high accuracy, fine-grained control, and the ability to handle occlusions, clutter, and dense instances~\citep{maninis2018deep}. However, the automated generation of point prompts using low-shot data, for personalized segmentation in dense scenarios, has largely remained unexplored. Recent works, such as SegGPT~\citep{wang2023seggpt}, PerSAM~\citep{zhang2023personalize} and Matcher~\citep{liu2023matcher}, introduce frameworks for one-shot personalized segmentation. Despite their effectiveness in sparsely populated scenes with clearly delineated objects, these methods show limited performance in dense scenarios (Fig.~\ref{fig:introfig}).

\begin{figure*}[t]
    \centering
    
    \includegraphics[width=1\linewidth]{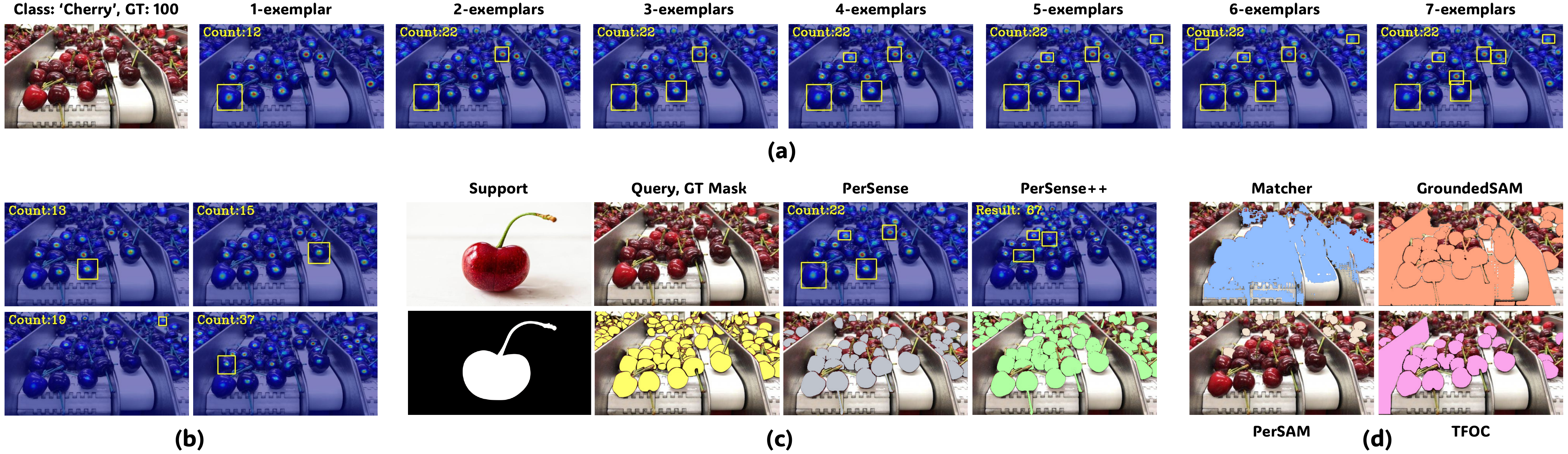}
    \caption{(a) DM accuracy in PerSense, driven by SAM score-based exemplar selection in feedback mechanism. Results indicate that selection based solely on SAM scores lacks diversity consideration, leading to DMG performance saturation despite the increase in the number of exemplars. (b) Sensitivity of DM quality to exemplar choice. (c) PerSense vs PerSense\texttt{++}: DM accuracy (predicted count vs GT count) and segmentation performance. (d) Performance of SOTA training-free approaches under similar settings.}
    \label{fig:motfig}
\end{figure*}

We approach this problem by exploring density estimation methods, which utilize density maps (DMs) to capture the spatial distribution of objects in dense scenes. While DMs effectively estimate global object counts, they struggle with precise instance-level localization~\citep{pelhan2024dave}. To this end, \textbf{\emph{we introduce PerSense}, an end-to-end, training-free and model-agnostic one-shot framework} (Fig.~\ref{fig:introfig}), wherein we first develop a new baseline capable of autonomously generating instance-level candidate point prompts via a proposed Instance Detection Module (IDM), which exploits DMs for precise localization. We generate DMs using a density map generator (DMG) which highlights spatial distribution of object of interest based on input exemplars. To allow automatic selection of effective exemplars for DMG, we automate the process via a class-label extractor (CLE) and a grounding detector. Second, we design a Point Prompt Selection Module (PPSM) to mitigate false positives within the candidate point prompts using an adaptive threshold and box-gating mechanism. Both IDM and PPSM are plug-and-play components, seamlessly integrating into PerSense. Lastly, we introduce a robust feedback mechanism, which automatically  refines the initial exemplar selection by identifying multiple rich exemplars for DMG based on the initial segmentation output of PerSense. This ability to segment personalized concepts in dense scenarios is pivotal for industrial automation tasks such as quality control and cargo monitoring using vision-based sensors. Beyond industry, it holds promise for medical applications, particularly in cellular-level segmentation. Extensive experiments demonstrate that PerSense outperforms SOTA methods in both performance and efficiency for dense settings. 

Finally, to our knowledge, no existing benchmark specifically targets segmentation in dense images. While datasets like COCO~\citep{lin2014microsoft}, LVIS~\citep{gupta2019lvis}, and FSS-1000~\citep{li2020fss} include multi-instance images, they lack dense scenarios due to limited object counts. For instance, LVIS averages only 3.3 instances per category. To bridge this gap, \textit{we introduce PerSense-D, a one-shot segmentation benchmark for dense images. It comprises 717 images across 28 object categories, with an average of 53 instances per image, along with dot annotations and ground truth masks}. Featuring heavy occlusion and clutter, PerSense-D offers both class-wise and density-based categorization for fine-grained evaluation in real-world dense settings.


This work is an extension of our previous work PerSense~\citep{siddiqui2025persense}. Through extensive analysis, we identify three key improvement areas in the PerSense pipeline: \textbf{(i)} PerSense employs a feedback mechanism for automated exemplar selection based on SAM scores from the initial segmentation. While this improves segmentation by enhancing DM accuracy, relying solely on SAM scores limits exemplar diversity and leads to premature saturation of DMG performance. As shown in Fig.\ref{fig:motfig}a, beyond the first few exemplars, object counts saturate, making additional exemplars redundant. Moreover, Fig.\ref{fig:motfig}b illustrates the sensitivity of DM to exemplar choice, establishing exemplar selection as a critical bottleneck. These findings motivate the need for a systematic strategy that promotes exemplar diversity and representativeness while preserving PerSense’s training-free nature. \textbf{(ii)} The IDM in PerSense employs a contour-based approach on DMs to extract candidate point prompts via centroid detection. However, in complex composite density regions, the contour-based approach may fail to adequately separate merged objects, causing missed detections of child instances. \textbf{(iii)} Although PPSM effectively filters out a majority of spurious prompts through its adaptive thresholding and box-gating mechanisms, a few false positives may still reach the decoder, triggering background or irrelevant segmentations that degrade performance.


To address these limitations, we propose PerSense\texttt{++}, an enhanced version of PerSense and make the following key contributions. \textbf{(1)} We introduce a diversity-aware exemplar selection strategy within the feedback mechanism. Beyond SAM scores, PerSense\texttt{++} incorporates feature diversity (semantic variation) and scale diversity (object size variation) to compute a weighted score, guiding the selection of diverse and representative exemplars. As shown in Fig.\ref{fig:motfig}c, this yields higher DM accuracy than PerSense while using fewer exemplars (3 vs. 4), leading to improved segmentation in dense scenes. For context, Fig.\ref{fig:motfig}d compares performance with SOTA methods. \textbf{(2)} We enhance the IDM pipeline with a peak detection stage that identifies local intensity maxima within the DM, complementing contour-based centroid extraction. Contour detection isolates merged regions without clear peaks, while peak detection resolves multiple instances within contiguous contours. Their union yields the final candidate point prompts. \textbf{(3)} We also introduce an Irrelevant Mask Rejection Module (IMRM), a lightweight post-processing block that discards any segmentation outliers (masks triggered by spurious prompts) deviating from the majority cluster of valid masks, ensuring only consistent target-class masks are retained. \textbf{(4)} To facilitate fine-grained analysis in complex scenes, we curate two new subsets, COCO-20\textsuperscript{d} and LVIS-92\textsuperscript{d}, comprising densely populated images with diverse object types, and demonstrate the generalization ability of our approach on these dense subsets. \textbf{(5)} Lastly, we evaluate object counting to assess DM accuracy, comparing PerSense with PerSense\texttt{++}, and provide further insights through extended experiments and ablations.

\begin{figure*}[t]
    \centering
    \includegraphics[width=1\linewidth]{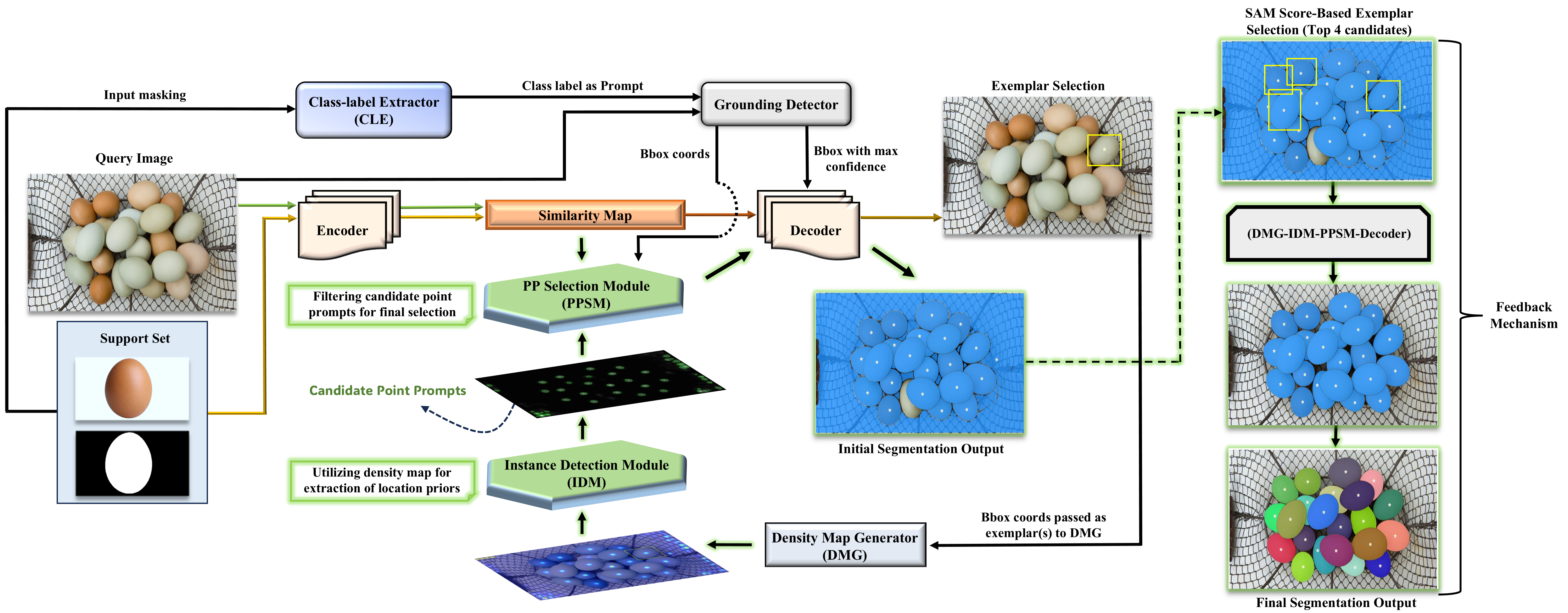}
    \caption{Overall architecture of our PerSense: an \textit{end-to-end}, \textit{training-free} and  \textit{model-agnostic} \textit{one-shot} framework. Support set guides exemplar selection in the query image for DM generation via DMG. The Instance Detection Module (IDM) extracts candidate point prompts leveraging DM, which are refined by the Point Prompt Selection Module (PPSM). A feedback mechanism further improves DM through SAM score-based exemplar selection, enabling IDM and PPSM to produce more accurate prompts for final segmentation.}
    \label{fig:mainfig}
\end{figure*}

\begin{figure}[t]
    \centering
    \includegraphics[width=1\linewidth]{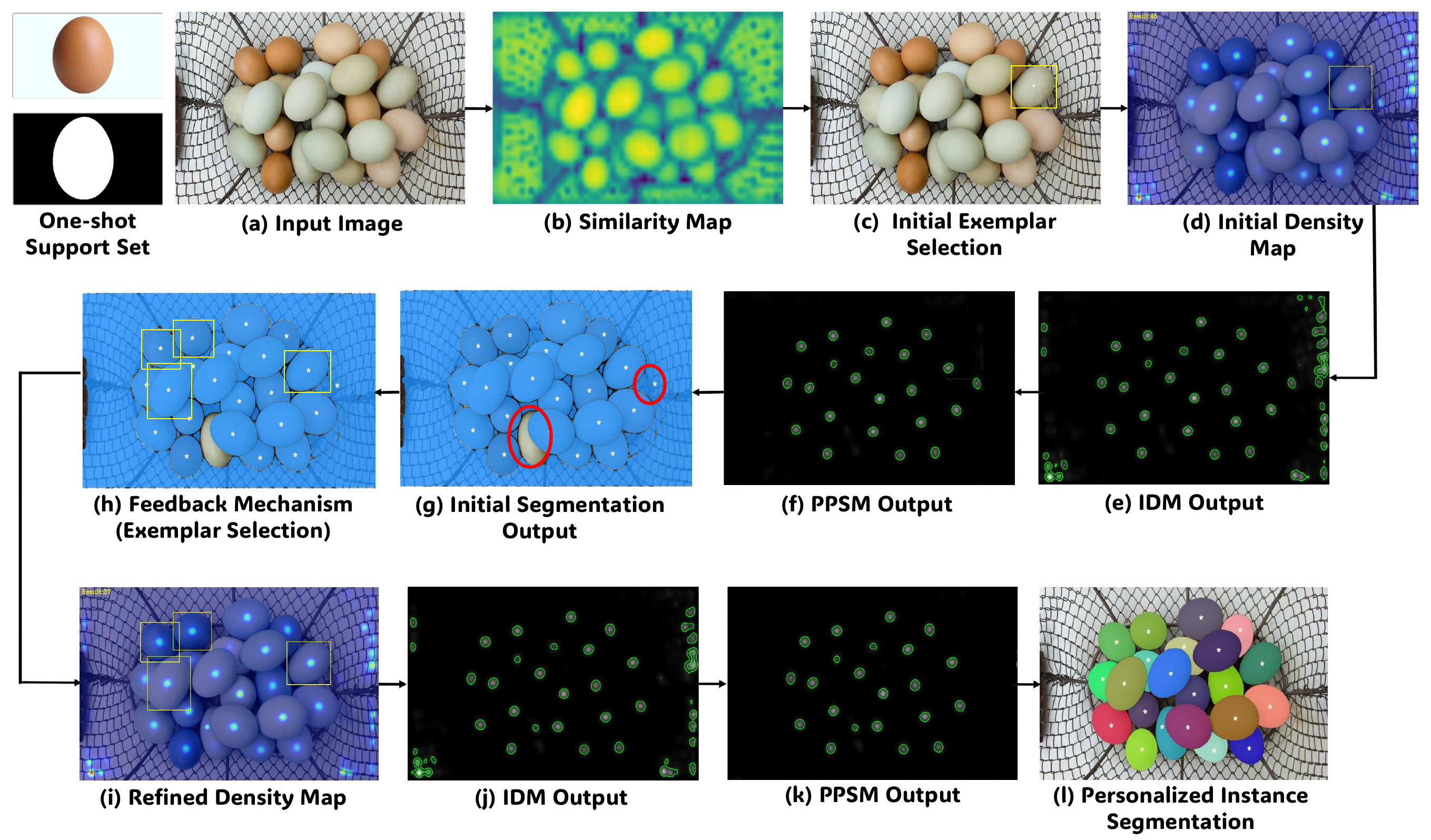}
    \caption{Step-by-step workflow of PerSense components. From an input image (a), a similarity map (b) is generated using the support set. Exemplar selection (c), guided by similarity scores and grounded detections, produces an initial density map (DM) (d) via DMG.  The IDM processes this DM to generate candidate point prompts (e), which are refined by PPSM (f) to filter false positives. The decoder yields an initial segmentation based on refined prompts (g); however, a few false positive prompts still remain alongside false negatives (red circle). The feedback mechanism (h) leverages initial segmentation to improve exemplar selection based on SAM scores, generating an improved DM (i). IDM and PPSM extracts refined prompts (j, k) from improved DM leading to the final segmentation output (l).}
    \label{fig:stepwise}
\end{figure}

\section{Related Work}\label{relatedwork}
\noindent\textbf{One-shot personalized segmentation:} As discussed in Sec.~\ref{intro}, SAM lacks semantic awareness, limiting its ability to segment personalized visual concepts. To address this, PerSAM~\citep{zhang2023personalize} introduces a training-free, one-shot segmentation framework using SAM, effectively segmenting multiple instances of the same category through an iterative masking approach. However, when applying PerSAM to dense images with many instances of the same object, several challenges may arise. (1) Its iterative masking becomes computationally expensive as the number of iterations scales with object count. (2) Confidence map accuracy degrades as more objects are masked, making it harder to distinguish overlapping instances. (3) PerSAM’s confidence thresholding strategy which halts the process when the confidence score drops below a set threshold, may lead to premature termination of segmentation process, even when valid objects are still present. In contrast, PerSense leverages DMs to generate precise instance-level point prompts in a single pass, eliminating the need for iterative masking and thereby improving segmentation efficiency in dense scenes. Another one-shot segmentation method, SLiMe~\citep{khani2023slime}, enables personalized segmentation based on segmentation granularity in the support set, rather than object category. Despite its strong performance, SLiMe tends to produce noisy segmentations for small objects due to the smaller attention maps extracted from Stable Diffusion~\citep{rombach2022high} compared to the input image.

Painter~\citep{wang2023images} introduces a unified vision model with in-context prompting, eliminating the need for downstream fine-tuning. However, its reliance on masked image modeling prioritizes coarse global features, limiting its ability to capture fine-grained details and handle dense scene complexity. SegGPT~\citep{wang2023seggpt} builds on Painter by introducing a random coloring approach for in-context training, improving generalization across segmentation tasks. However, in one-shot settings, this scheme may oversimplify densely packed regions, making it difficult to distinguish overlapping objects. Matcher~\citep{liu2023matcher} integrates a feature extraction model with a class-agnostic segmentation model, using bidirectional matching for semantic alignment. However, its instance-level matching is limited, impacting instance segmentation performance. Matcher relies on reverse matching to remove outliers and $k$-means clustering for instance sampling, which can become a bottleneck in dense scenes due to varying object scales. In contrast, PerSense leverages DMG to generate personalized DMs, eliminating the need for clustering and sampling. Additionally, Matcher forwards the bounding box of the matched region as a box prompt to SAM, which inherits limitations of box-based detections, particularly in dense environments.

\noindent\textbf{Interactive segmentation:} Recently, interactive segmentation has gained attention, with models like InterFormer~\citep{Huang_2023_ICCV}, MIS~\citep{Li_2023_ICCV}, and SEEM~\citep{zou2024segment} offering user-friendly interfaces but relying on manual input from the user, which limits scalability. More recently,  Semantic-SAM~\citep{li2024segment} improves upon vanilla SAM by incorporating semantic-awareness and multi-granularity segmentation, however it still requires manual prompts and does not explicitly generalize from one-shot reference to all instances of the same category within the image. In contrast, PerSense automates instance-specific point prompt generation from one-shot data, enabling personalized segmentation without manual intervention.

\section{PerSense}\label{sec3}

We introduce PerSense, a training-free and model-agnostic one-shot framework designed for personalized instance segmentation in dense images (Fig.~\ref{fig:mainfig}). In following sections, we detail the core components of our PerSense framework. The step-by-step working of PerSense components is depicted in Fig.~\ref{fig:stepwise}.


\begin{figure*}[!t]
    \centering
    \includegraphics[width=1\linewidth]{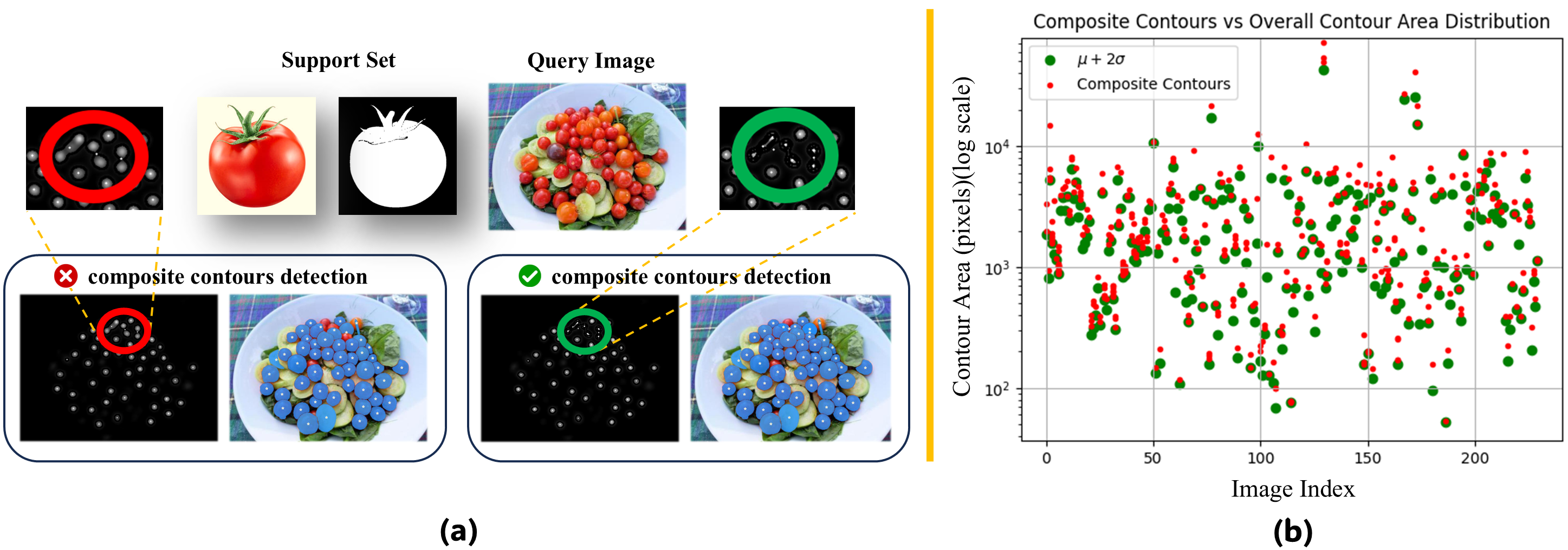}
    \caption{(a) Without identifying composite contours, multiple object instances may be incorrectly grouped (red circle). Identification of composite contours (green circle) enables accurate localization of child contours. (b) The plot illustrates the presence of composite contours beyond \( \mu + 2\sigma \) in the contour area distribution for 250 dense images. (best viewed in zoom)}
    \label{fig:comp_contour} \vspace{-1em}
\end{figure*}

\subsection{Class-label Extraction and Exemplar Selection}
\label{subsection:class-label-extraction}

PerSense operates as a one-shot framework, leveraging a support set to guide personalized segmentation in a query image based on semantic similarity with the support object. First, input masking is applied to the support image using a coarse support mask to isolate the object of interest. The masked image is then processed by the CLE with the prompt, "\textit{Name the object in the image?}". The CLE generates a description, from which the noun is extracted as the object category. This category serves as a prompt for the grounding detector, enabling personalized object detection in the query image. To refine the prompt, the term "all" is prefixed to the class label. Next, we compute the cosine similarity score \( S_{score} \) between query \( Q \) and support \( S_{supp} \) features, extracted by the encoder:
\begin{equation}
S_{score}(Q, S_{supp}) = \text{sim}(f(Q), f(S_{supp})),
\end{equation}
where \( f(\cdot) \) represents the encoder. Utilizing this score along with detections from the grounding detector, we extract the positive location prior. Specifically, we identify the bounding box \( B_{max} \) with the highest detection confidence and proceed to locate the pixel-precise point \( P_{max} \) with the maximum similarity score within this bounding box:
\begin{equation}
P_{max} = \arg\max_{P \in B_{max}} S_{score}(P, S_{supp}),
\end{equation}
where \( P \) represents candidate points within the bounding box \( B_{max} \). This identified point serves as the positive location prior, which is subsequently fed to the decoder for segmentation. Additionally, we extract the bounding box surrounding the segmentation mask of the object. This bounding box is then forwarded as an exemplar to the DMG for generation of DM.

\subsection{Instance Detection Module (IDM)}
\label{subsection:IDM}

The IDM begins by converting the DM from the DMG into a grayscale image \( I_{gray} \). A binary image \( I_{binary} \) is created from \( I_{gray} \) using a pixel-level threshold \( T \) (\( T \in [0, 255] \)):
\vspace{-3pt}
\begin{equation}
I_{binary}(x, y) = 
\begin{cases} 
1 & \text{if } I_{gray}(x, y) \geq T \\
0 & \text{if } I_{gray}(x, y) < T 
\end{cases}
\end{equation}
A morphological erosion operation is then applied to \( I_{binary} \) using a \( 3 \times 3 \) kernel \( K \):
\begin{equation}
I_{eroded}(x, y) = \min_{(i,j) \in K} I_{binary}(x+i, y+j),
\end{equation}
where \( I_{eroded} \) is the eroded image, and \( (i,j) \) iterates over the kernel \( K \) to refine the boundaries and eliminate noise from the binary image. We deliberately used a small kernel to avoid damaging the original densities of true positives. Next, contours are extracted from \( I_{eroded} \), and their areas $A_\text{ctr}$ are modeled as a Gaussian distribution:
\begin{equation}
A_\text{ctr} \sim \mathcal{N}(\mu, \sigma^2)
\end{equation}
where $\mu$ represents the mean contour area, corresponding to the typical object size, and $\sigma$ denotes the standard deviation, capturing variations in contour areas due to differences in object sizes among instances.
The mean and standard deviation are computed as:
\begin{equation}
\mu = \frac{1}{N} \sum_{i=1}^N A_i, \quad \sigma = \sqrt{\frac{1}{N} \sum_{i=1}^N (A_i - \mu)^2}
\end{equation}
where $N$ is the number of detected contours. Composite contours, which encapsulate multiple objects within a single contour, are identified using a threshold \( T_\text{comp} \), defined as \( \mu + 2\sigma \) based on the contour size distribution (Fig.~\ref{fig:comp_contour}). These regions, though rare, are detected as outliers exceeding \( T_\text{comp} \). The probability of a contour being composite is given by:



\vspace{-5pt}
\begin{equation}
P(A_\text{ctr} > T_\text{comp}) = 1 - \Phi\left(\frac{T_\text{comp} - \mu}{\sigma}\right)
\end{equation}
where $\Phi$ is the cumulative distribution function (CDF) of the standard normal distribution. For each composite contour, a distance transform $D_\text{transform}$ is applied to reveal internal sub-regions representing individual object instances:
\begin{equation}
D_\text{transform}(x, y) = \min_{(i, j) \in B} \| (x, y) - (i, j) \|
\end{equation}
where $B$ represents contour boundary pixels and $(x, y)$ are the coordinates of each pixel within the region of interest. A binary threshold applied to $D_\text{transform}$ segments sub-regions within each composite contour, enabling separate identification of overlapping objects in dense scenarios. For each detected contour (parent and child), the centroid is calculated using spatial moments:
\begin{equation}
cX = \frac{M_{10}}{M_{00} + \epsilon}, \quad cY = \frac{M_{01}}{M_{00} + \epsilon}
\end{equation}
where $M_{pq}$ are the spatial moments of the contour, and $\epsilon$ is a small constant to prevent division by zero. These moments are computed as:

\begin{equation}
(M_{00}, M_{10}, M_{01}) = \sum_x \sum_y I(x, y) .\big(1, x, y\big)
\end{equation}
\noindent where $I(x, y)$ is the pixel intensity at position $(x, y)$. These centroids serve as candidate point prompts, accurately marking the locations of individual object instances in dense scenarios. The candidate point prompts are subsequently forwarded to PPSM for final selection.

\begin{figure}[!t]
    \centering
    \includegraphics[width=0.9\linewidth]{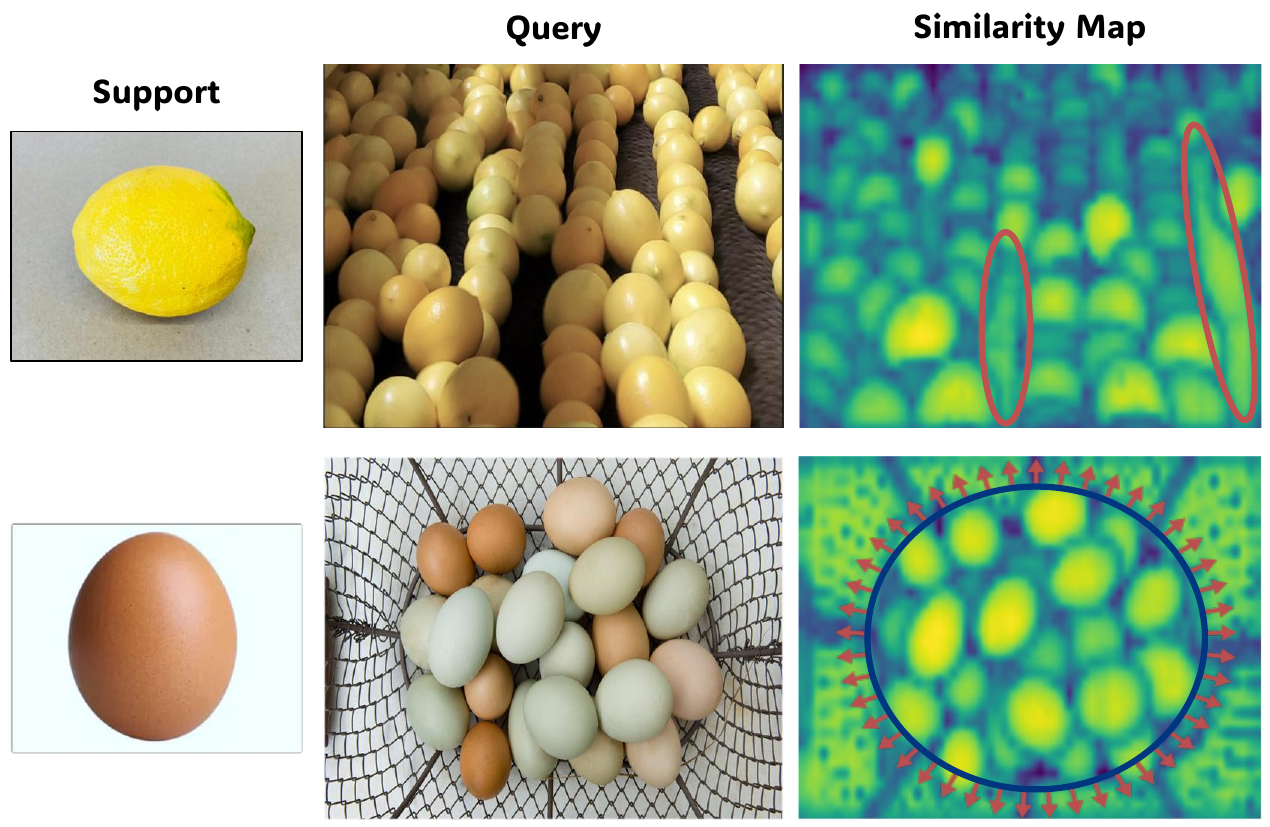}
    \caption{Cosine similarity captures directional alignment in feature space instead of true semantics, leading to high similarity scores in irrelevant and background regions (highlighted in brown).}
    \label{fig:ppsm2}
\end{figure}

\subsection{Point Prompt Selection Module (PPSM)}
\label{subsection:PPSM}

The PPSM serves as a critical component in the PerSense pipeline, filtering candidate point prompts before forwarding the selected points to the decoder for segmentation. Each candidate point from IDM is evaluated based on its query-support similarity score, using an adaptive threshold that adjusts dynamically to object density. This ensures a balance between true positive inclusion and false positive elimination in dense scenes.
We statistically model the adaptive threshold in PPSM, scaling it based on object count. 
Assuming similarity score $S(x, y)$  approximate a Gaussian distribution with mean $\mu$ and variance $\sigma^2$, we consider the maximum score $S_\text{max}$ as the upper bound, representing the most aligned point with the target feature. The adaptive threshold \( T_\text{adapt} \) for point selection is then defined as:




\begin{figure*}[t]
    \centering
    \includegraphics[width=1\linewidth]{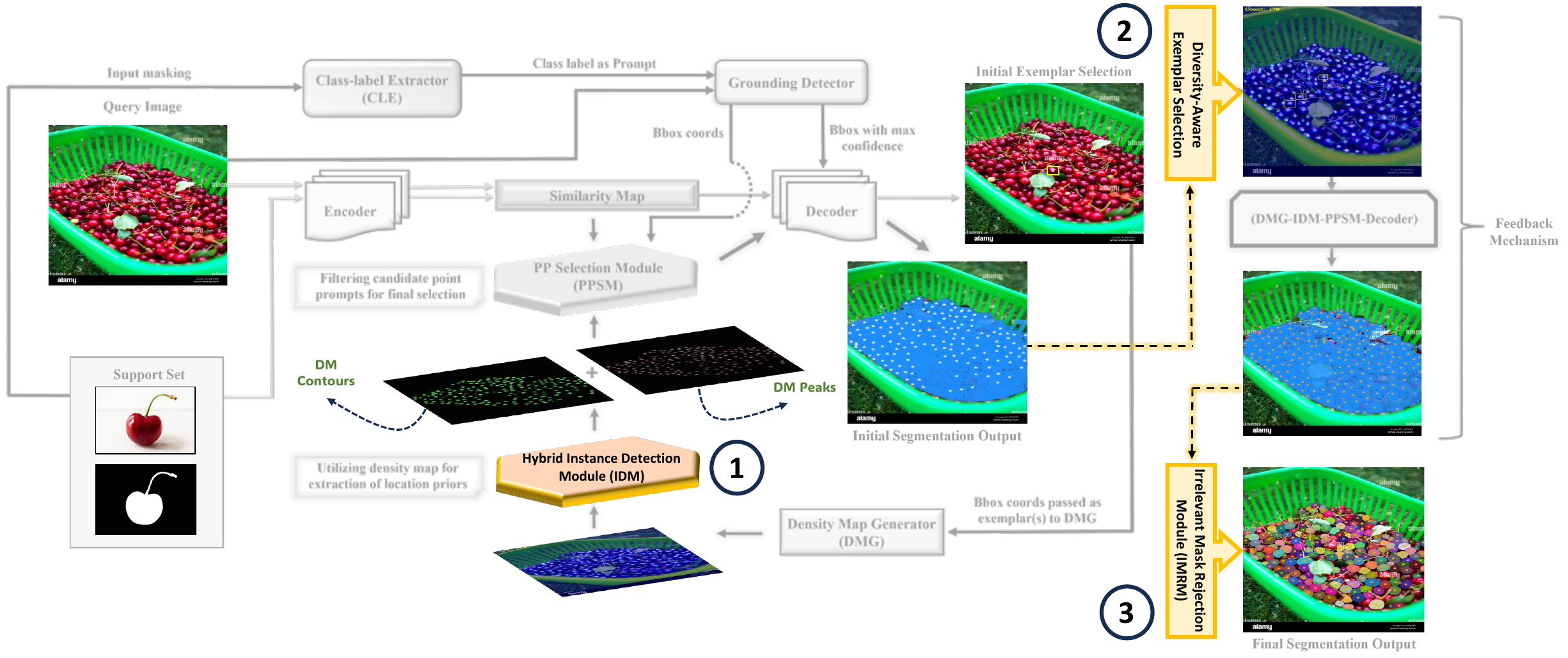}
\caption{Overview of the PerSense\texttt{++} framework. Colored regions highlight the enhancements introduced in PerSense\texttt{++}, while greyed-out components represent the original PerSense architecture. PerSense\texttt{++} introduces three key improvements: (1) a hybrid Instance Detection Module (IDM) that incorporates peak detection alongside contour-based methods to generate candidate point prompts from DMs; (2) a diversity-aware exemplar selection strategy that extends beyond SAM scores by incorporating feature and scale diversity to compute a weighted score for selecting representative exemplars; and (3) an Irrelevant Mask Rejection Module (IMRM) that filters out outlier segmentation masks by identifying those that deviate significantly from the majority cluster of valid object instances.}

    \label{fig:mainfig_persesepp}
\end{figure*}

\begin{equation}
T_\text{adapt} = \frac{S_{\text{max}}}{C / k}, \quad \text{for } C > 1
\end{equation}
where $C$ and $k$ represent the object count and normalization constant, respectively.  When $C = 1$, the point associated with $S_\text{max}$ is selected as the prompt. We choose \( k = \sqrt{2} \), based on empirical results presented in Sec.~\ref{ablations}. This naturally raises the question: why not estimate the spread of the similarity scores and define the threshold based on the mean and variance of the score distribution? The key challenge lies in the nature of cosine similarity, which measures directional alignment in feature space rather than true semantic correspondence. As a result, high similarity scores can arise in irrelevant or background regions (Fig.~\ref{fig:ppsm2}), introducing noise that may artificially inflate the estimated mean and variance of the score distribution. Applying a threshold based on these inflated statistics would likely set the threshold high, causing true positives to be incorrectly rejected. In contrast, our adaptive thresholding strategy modulates the threshold directly based on the scene's density level, ensuring tighter thresholds in sparse scenes (to minimize false positives) and more flexible thresholds in dense scenes (to avoid missing true positives). As \( T_\text{adapt} \) becomes more permissive with increasing \( C \), the probability of candidate points surpassing the threshold also rises, ensuring greater flexibility in dense scenarios. The probability $P$ of a randomly selected point having a similarity score $S$ above \( T_\text{adapt} \) is given by:

\begin{equation}
P(S \geq T_\text{adapt}) = 1 - \Phi\left(\frac{T_\text{adapt} - \mu}{\sigma}\right)
\end{equation}
where $\Phi$ is the CDF of the standard normal distribution. Substituting for \( T_\text{adapt} \), we get:
\begin{equation}
P(S \geq T_\text{adapt}) = 1 - \Phi\left(\frac{\frac{S_{\text{max}}}{C / k} - \mu}{\sigma}\right)
\end{equation}
\vspace{-1pt}This density-aware adjustment mitigates the risk of inadvertently excluding true positives, a limitation commonly observed with fixed thresholds tuned primarily to suppress false positives. Such adaptability is crucial as query-support similarity scores vary significantly, even with minor intra-class differences. In highly dense images (\( C > 50 \)), the score distribution widens due to increased intra-class variability, making dynamic thresholding essential for robust and reliable prompt selection. Our adaptive threshold is intentionally designed to adopt lower values for highly dense images, which is crucial due to the nature of cosine similarity. This prevents the inadvertent rejection of true positives that may have moderate similarity scores. However, such a permissive threshold also increases the risk of false positives from IDM passing through to the decoder, potentially introducing segmentation noise in dense scenes. To address this, we introduce a box gating mechanism that enhances the spatial precision of the selected prompts. This mechanism leverages bounding box detections from grounding detector to constrain the spatial validity of candidate points. In short, a point is retained only if (i) its similarity score exceeds \( T_\text{adapt} \), and (ii) it falls within at least one of the boxes predicted by the grounding detector. This ensures selected prompts are both semantically aligned and spatially grounded.

\begin{figure*}[!t]
    \centering
    \includegraphics[width=1\linewidth]{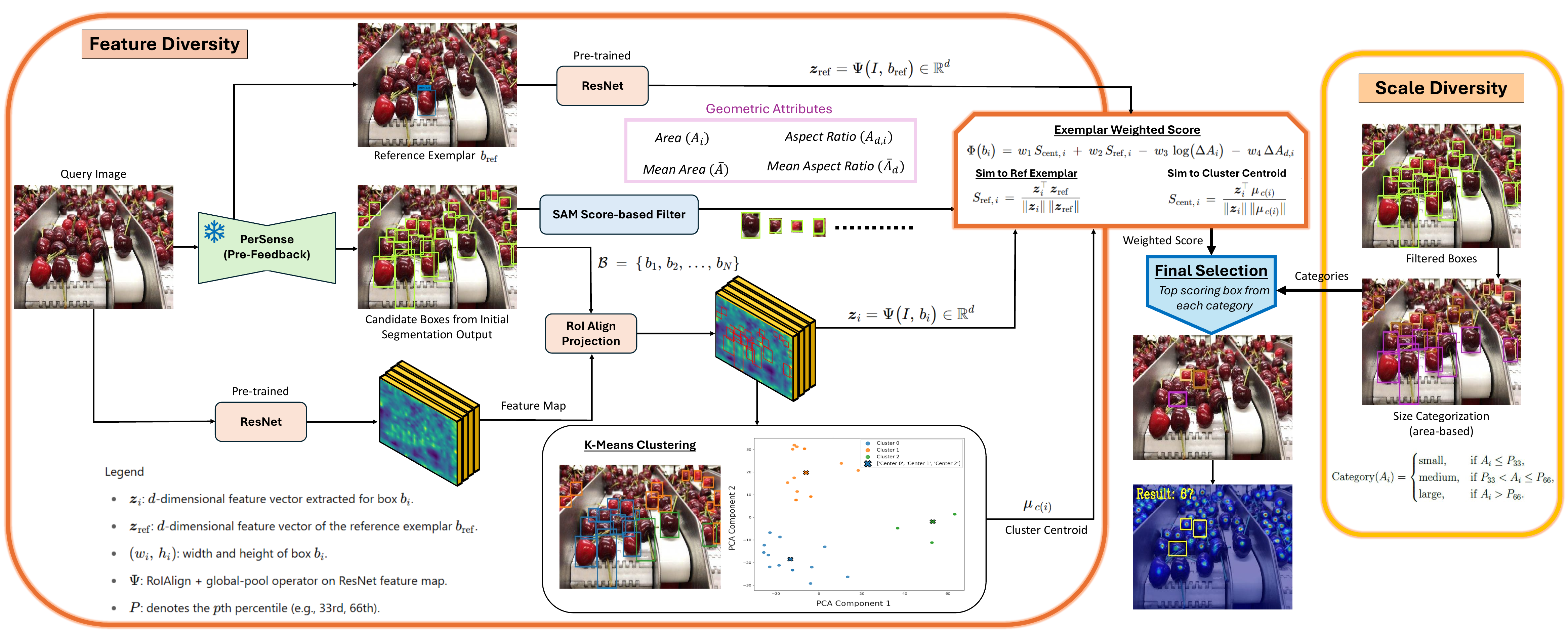}
    \caption{Overview of the proposed diversity-aware exemplar selection strategy in PerSense\texttt{++}. Initial segmentation from PerSense (pre-feedback) provides candidate exemplars, filtered by SAM score. RoIAlign extracts feature vectors from bounding boxes, which are clustered via $k$-means to promote feature diversity. A weighted score, combining similarity to cluster centroids and reference exemplar along with deviation from mean geometric attributes, ranks candidates. To ensure scale diversity, candidates are binned by size using percentile thresholds, and top-scoring exemplars from each bin are selected. Resulting exemplars are semantically and spatially diverse, improving DM quality and segmentation accuracy.}
    \label{fig:persenseplus_exemplar_selection} 
\end{figure*}

\subsection{Feedback Mechanism}
\label{subsection:Feedback_mechanism}

PerSense proposes a feedback mechanism to enhance the exemplar selection process for the DMG by leveraging the initial segmentation mask (\( M_{seg} \)) from the decoder and the corresponding SAM-generated mask scores (\( S_{mask} \)).
\begin{equation}
C_{Top} = \text{Top}_m(M_{seg}, S_{mask}, k),
\end{equation}
where \( C_{Top} \) represents the set of the top \( m \) candidates, selected based on their mask scores. In our case  \( m = 4 \)  (see Sec.~\ref{ablations}). These candidates are then forwarded to DMG in a feedback manner, enhancing the quality of the DM and, consequently, the segmentation performance. The quantitative analysis of this aspect is further discussed in Sec.~\ref{subsection:persenseresults}, which explicitly highlights the value added by the proposed feedback mechanism.

    


\section{PerSense\texttt{++}}
\label{persense++}

While PerSense laid the foundation for training-free personalized segmentation in dense scenes, our analysis reveals three critical limitations that restrict its performance. First, exemplar selection based solely on SAM scores lacks diversity awareness, leading to redundant exemplars and early saturation of DM accuracy. Second, the contour-based instance detection in IDM struggles to separate merged objects in complex density patterns, resulting in missed instances. Third, despite PPSM’s filtering mechanisms, a few false-positive prompts can still reach the decoder and trigger irrelevant segmentations. To overcome these challenges, we introduce PerSense\texttt{++} (Fig.~\ref{fig:mainfig_persesepp}), an enhanced framework that incorporates: (i) a diversity-aware exemplar selection strategy combining semantic and scale variations, (ii) a hybrid instance detection mechanism in IDM that fuses contour and peak-based cues, and (iii) a lightweight  Irrelevant Mask Rejection Module (IMRM) to discard inconsistent masks in a post-processing step. These improvements collectively enhance segmentation accuracy, robustness, and efficiency, while preserving PerSense’s core advantage of being entirely training-free. Further details are presented in the following subsections.


\subsection{Diversity-Aware Exemplar Selection}
\label{diversityawareexemplar}

The PerSense feedback mechanism, as described in Sec.~\ref{subsection:Feedback_mechanism}, selects exemplars for the DMG based solely on SAM scores. However, as discussed in Sec.~\ref{intro}, this selection strategy exhibits diminishing returns, with DM accuracy saturating after the first few exemplars (see Fig.~\ref{fig:motfig}a), indicating that additional exemplars contribute little to no new information about the target object. While the SAM score reflects the reliability of the segmentation output, it does not capture feature diversity (semantic variation) or scale diversity (variation in object size) across exemplars. As a result, semantically similar and similarly sized exemplars may be repeatedly selected, introducing redundancy into the exemplar set. Given that DMGs are highly sensitive to exemplar selection, an effect illustrated in Fig.~\ref{fig:motfig}b, this highlights the need for a diversity-aware exemplar selection strategy within the feedback mechanism. Incorporating semantic and scale diversity ensures that the selected exemplars better represent the range of variations in the object of interest, ultimately improving DM quality and segmentation performance.

As discussed in Sec.~\ref{subsection:Feedback_mechanism} and illustrated in Fig.~\ref{fig:mainfig} and Fig.~\ref{fig:stepwise}, the initial segmentation output from the decoder serves as the input to the feedback mechanism for automatic exemplar selection. This output includes segmentation masks, each accompanied by its respective SAM score, to facilitate exemplar selection. Bounding boxes that encompass the boundaries of each individual mask are utilized as candidate exemplars in the selection process. PerSense\texttt{++} aims to enhance this process by integrating \textit{feature diversity} and \textit{scale diversity} into the exemplar selection criterion, alongside SAM score, ensuring a more robust and representative set of selected exemplars. Fig.~\ref{fig:persenseplus_exemplar_selection} highlights the core steps involved in the proposed exemplar selection strategy.

\begin{figure}[t]
    \centering
    \includegraphics[width=1\linewidth]{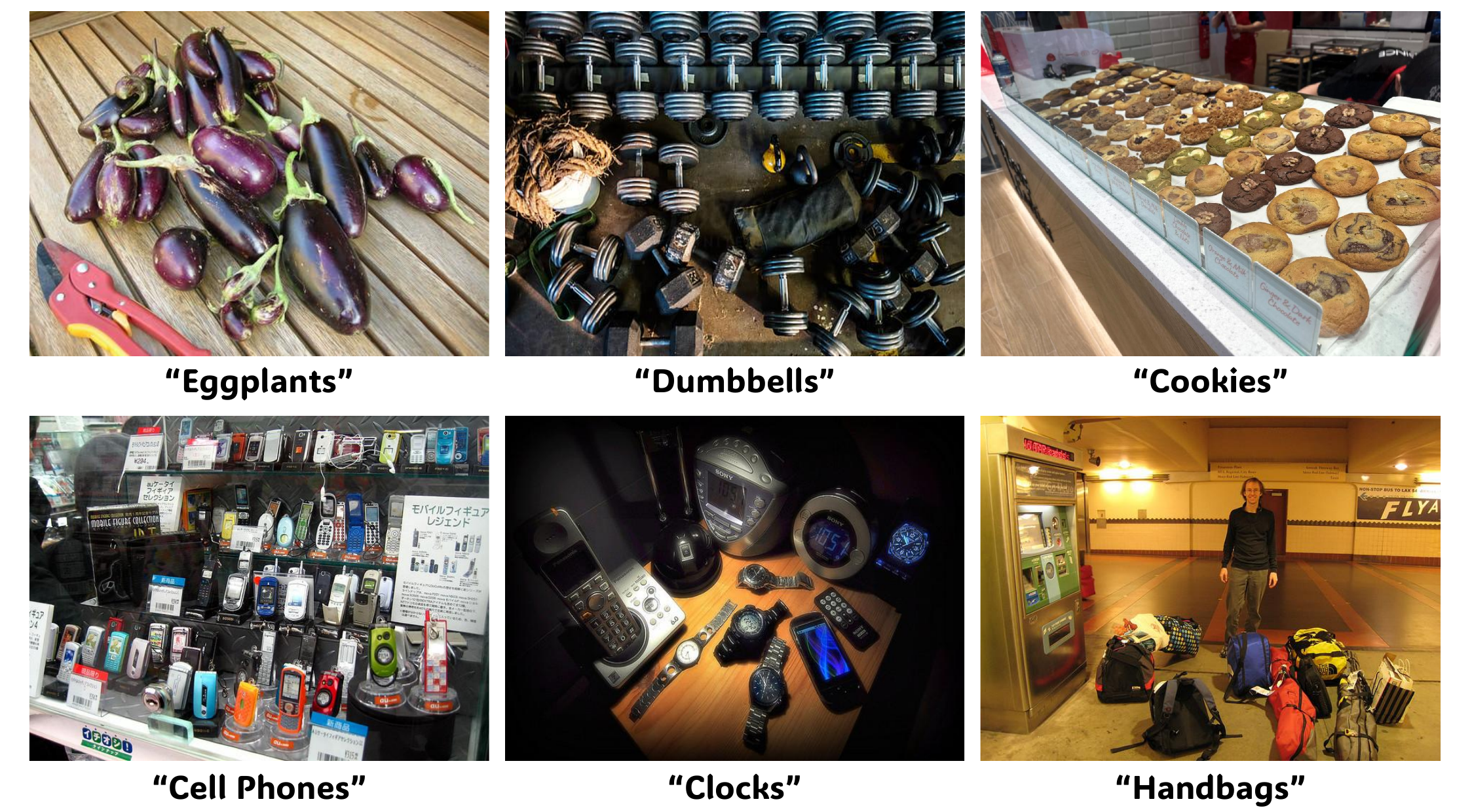}
    \caption{Illustration of intra-class variations in different classes, highlighting variation in texture, shape, and structural features within the same class across the same image.}
    \label{fig:intraclassvariations}
\end{figure}

\noindent\textbf{Feature Diversity: }The introduction of feature diversity is intended to capture intra-class variations within the object of interest, ensuring that the exemplar set selected for the DMG effectively represents these variations. As illustrated in Fig.~\ref{fig:intraclassvariations}, certain classes exhibit pronounced differences in texture, shape, and structural characteristics across instances within the same image. The key steps of the proposed feature diversity pipeline are discussed below.


\textit{Feature Extraction:} We utilize ResNet50~\citep{he2016deep} pre-trained on ImageNet~\citep{deng2009imagenet} to extract high-dimensional feature representations for all candidate exemplars. To preserve the independence of the exemplar selection process from the PerSense\texttt{++} architecture, we deliberately refrain from using the PerSense\texttt{++} encoder, rendering this module plug-and-play and compatible with segmentation outputs from any network. Rather than directly cropping each bounding box $b_i$ and resizing it to a fixed size (e.g., $224 \times 224$ in ResNet50), which can distort small objects and degrade feature quality, we instead extract a global feature map from the full image using ResNet50. We then apply \textit{RoIAlign} to project each bounding box $b_i \in B$ onto this feature map to obtain a semantically meaningful and spatially precise feature vector:
\begin{equation}
\mathbf{z}_i = \Psi(I, b_i) \in \mathbb{R}^d
\end{equation}
where $\Psi$ denotes the RoIAlign operation applied on the ResNet feature map of image $I$, and $\mathbf{z}_i$ represents the feature vector for box $b_i$. This approach is particularly crucial for preserving fine-grained details in small regions, which would otherwise be lost if resized to larger fixed dimensions.

\textit{Feature Normalization:} To ensure consistency and comparability across features, we normalize the feature vectors using standard normalization, i.e., \( \tilde{\mathbf{z}}_i = (\mathbf{z}_i - \mu)/\sigma \),
where $\mu$ and $\sigma$ are the mean and standard deviation computed over all candidate feature vectors $\{\mathbf{z}_i\}_{i=1}^N$.

\textit{Feature-based Clustering:} We employ $k$-means clustering algorithm~\citep{caron2018deep} to group the normalized feature vectors $\{\tilde{\mathbf{z}}_i\}_{i=1}^N$ into $k$ clusters based on their semantic similarity. The clustering objective is:
\begin{equation}
\min_{\{\mu_{c(i)}\}, \{z_{ij}\}} \sum_{j=1}^k \sum_{i=1}^N z_{ij} \left\|\tilde{\mathbf{z}}_i - \mu_{c(i)}\right\|^2
\end{equation}
where $\mu_{c(i)}$ is the centroid of the cluster assigned to box $i$, $z_{ij} \in \{0, 1\}$ indicates assignment of box $i$ to cluster $j$, and $k$ represents number of clusters. The cluster centroid $\mu_{c(i)}$ represents the mean feature vector of the cluster and serves as a semantic anchor in the feature space. A straightforward strategy could involve selecting the candidate box closest to the cluster centroid as the exemplar. However, this presents two critical challenges:

First, the candidate set includes masks derived from the initial segmentation output of PerSense that may be noisy, incomplete, or inaccurate, which can lead to suboptimal exemplar selection and degraded DM quality. Although SAM scores, indicating the quality of each segmentation mask, are available for all candidate boxes, we deliberately avoid filtering by SAM score prior to clustering. This design choice ensures that potentially informative features are not prematurely excluded, as even partial or incomplete masks are likely to correspond to the target object and may capture useful semantic cues. However, the presence of such incomplete masks can bias the selection of exemplars, e.g., the bounding box closest to the cluster centroid may encompass only a portion of object, reducing its representativeness. To mitigate this, we apply SAM score-based filtering after performing clustering over the complete set of candidate exemplars. This allows cluster centroids to be computed from the full distribution of candidate features, while ensuring that the final exemplar selection is restricted to boxes with high-quality, reliable masks.

Second, while clustering promotes \textit{semantic (feature) diversity}, it does not account for \textit{scale diversity}, i.e., the variation in object sizes within an image. As illustrated in Fig.~\ref{fig:intraclassvariations}, a query image may contain instances of the same class at multiple scales (e.g., small and large eggplants). Relying solely on feature similarity may lead to the selection of exemplars with diverse appearance but similar scale, thereby failing to fully capture intra-class variation. Addressing this requires explicit incorporation of geometric attributes, which we integrate into our exemplar scoring and selection process as discussed in the following sections.

\begin{figure}[t]
    \centering
    \includegraphics[width=0.95\linewidth]{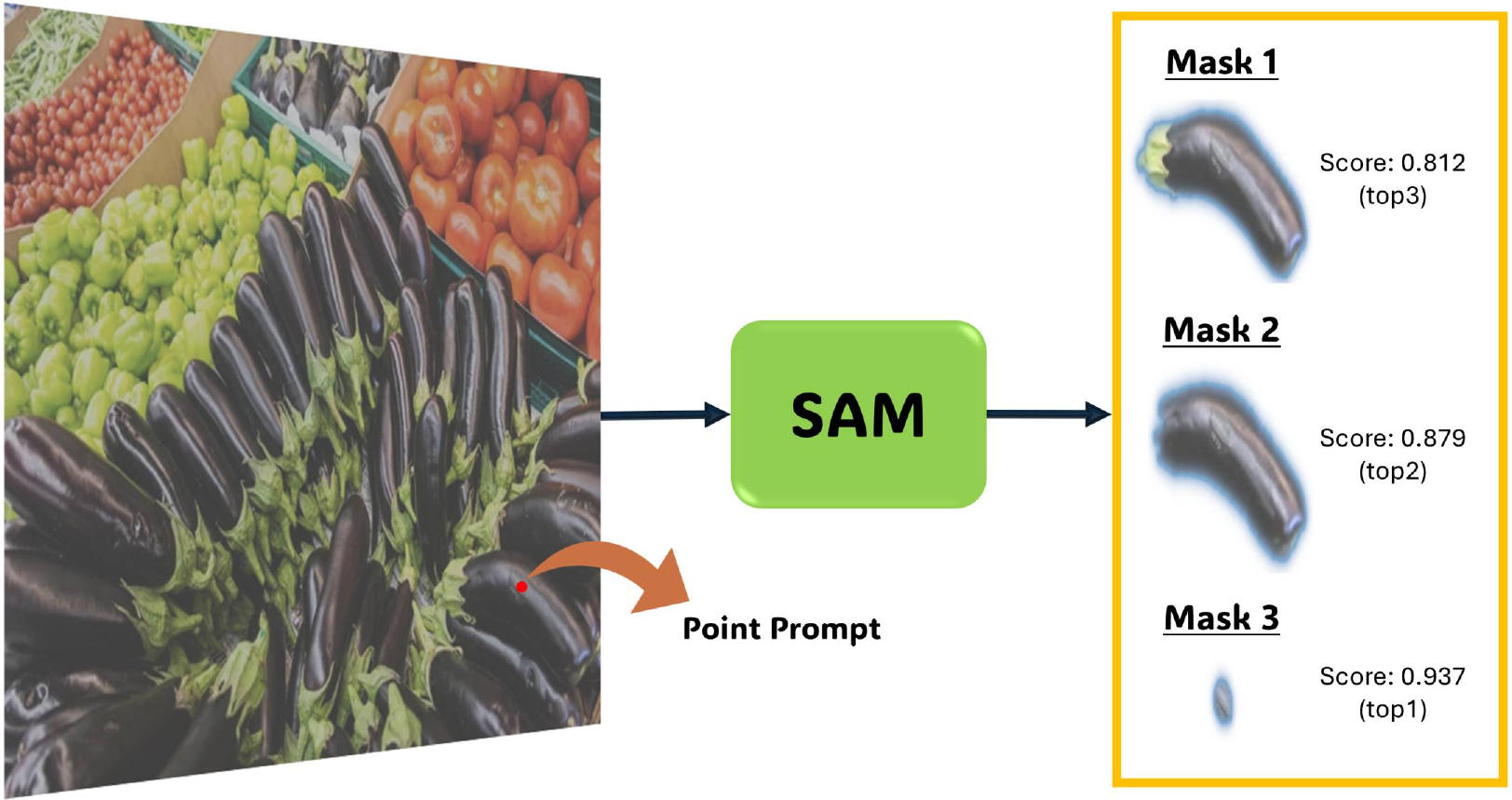}
    \caption[SAM limitation: Assigning high scores to masks of irrelevant regions]{The figure highlights SAM's limitation, demonstrating how it assigns high scores to segmentation masks of irrelevant regions or parts of the object due to its inherent semantic ambiguity. In this case, the goal was to segment the eggplant based on the given point prompt, but SAM's multimask output erroneously assigned the highest score to the mask corresponding to the sticker on the eggplant. }
    \label{fig:semanticambiguity}
\end{figure}


\noindent\textbf{SAM Score-based Filtration: }To ensure the quality of candidate exemplars, we perform a filtration step based on SAM scores. Let \( \mathcal{E} = \{e_1, e_2, \dots, e_n\} \) denote the set of candidate exemplars, and let \( \text{SAM}(e_i) \) represent the SAM score associated with exemplar \( e_i \). We define a fixed threshold \( T_{\text{SAM}} = 0.8 \), and retain only those exemplars that satisfy the following condition:
\begin{equation}
\mathcal{E}_{\text{filtered}} = \{e_i \in \mathcal{E} \;|\; \text{SAM}(e_i) \geq T_{\text{SAM}}\}
\end{equation}

\noindent Exemplars for which \( \text{SAM}(e_i) < T_{\text{SAM}} \) are discarded. The resulting filtered set \( \mathcal{E}_{\text{filtered}} \) serves as the input to subsequent stages of the exemplar selection process, ensuring that only high-quality exemplars are considered for final selection.

Given that the remaining candidates are now filtered based on segmentation quality, we proceed to select the final exemplars by identifying the exemplar from each cluster that lies closest to the cluster centroid in the feature space. However, a key challenge still remains. SAM-score-based filtering, although effective in promoting mask quality, suffers from a fundamental limitation: SAM is a class-agnostic segmentation model and lacks semantic meaning. As a result, it evaluates the quality of the generated mask purely based on alignment with image edges and objectness, without the consideration of semantic relevance to the intended object category. This can lead to undesired outcomes where high SAM scores are assigned to masks corresponding to irrelevant or incorrect regions. For example, as illustrated in Fig.~\ref{fig:semanticambiguity}, when the intended target is an eggplant, SAM’s multimask output, prompted by a point, assigns the highest score to a mask covering a sticker on the eggplant rather than the eggplant itself. This demonstrates a form of semantic ambiguity where, although the mask is high quality in terms of shape and boundaries, it does not semantically correspond to the object of interest. Consequently, relying solely on SAM score for exemplar filtering can lead to poor exemplar choices and degrade downstream segmentation performance in PerSense\texttt{++}.

\noindent\textbf{Weighted Score: }To refine exemplar selection and suppress irrelevant masks with high SAM scores, we identify a \textit{reference exemplar} within the query image to guide the process. We use the initial exemplar derived from the positive location prior as the reference exemplar (see Sec.~\ref{subsection:class-label-extraction}). A natural question arises: why not use the support set exemplar as the reference? The support set provides a category-level representation, which may not reflect the specific visual and contextual characteristics of the query image. Thus, a query-specific reference exemplar is more appropriate for precise alignment.

Selecting exemplars solely based on proximity to the cluster centroid may lead to the inclusion of boxes that, while well-embedded within the cluster, correspond only to parts of the object. This necessitates the inclusion of \textit{geometric attributes} to complement feature similarity and ensure accurate representation of the object of interest. Specifically, we utilize the mean area and mean aspect ratio of all boxes in the filtered set \( \mathcal{E}_{\text{filtered}} \) as reference geometric attributes. We define a \textit{weighted score} \( \Phi(b_i) \) that integrates semantic and geometric cues. For feature similarity, we consider both the similarity to the \textit{cluster centroid} and the \textit{reference exemplar}. This dual perspective ensures that if the cluster centroid diverges from the true object representation, the reference exemplar steers the selection toward relevant instances. The geometric terms include deviations in area and aspect ratio from the mean values, enforcing spatial consistency. The final score is computed as:

\begin{equation*}
\Phi(b_i) = w_1 S_{\text{cent}, i} + w_2 S_{\text{ref}, i} - w_3 \log(\Delta A_i) - w_4 \Delta A_{d, i}
\end{equation*}

\noindent where:
\begin{itemize}
    \item \( S_{\text{cent}, i} = \frac{\mathbf{z}_i^\top \mu_{c(i)}}{\|\mathbf{z}_i\|\|\mu_{c(i)}\|} \) is the cosine similarity between \( b_i \)'s feature vector \( \mathbf{z}_i \) and its cluster centroid \( \mu_{c(i)} \in \mathbb{R}^d \),
    \item \( S_{\text{ref}, i} = \frac{\mathbf{z}_i^\top \mathbf{z}_{\text{ref}}}{\|\mathbf{z}_i\| \|\mathbf{z}_{\text{ref}}\|} \) is the cosine similarity with the reference exemplar,
    \item \( \Delta A_i = |A_i - \bar{A}| \) is the absolute area deviation from the mean area \( \bar{A} \),
    \item \( \Delta A_{d, i} = |A_{d,i} - \bar{A}_d| \) is the absolute deviation in aspect ratio from the mean aspect ratio \( \bar{A}_d \),
    \item \( w_1, w_2, w_3, w_4 \) are scalar weights controlling the influence of each term. \((+)\) indicates "higher is better," while \((-)\) denotes "lower is better.
\end{itemize}

Since object areas can vary significantly, particularly in images with objects of diverse scales, the logarithmic transformation is applied to prevent area differences from disproportionately influencing the weighted score. This formulation ensures that the final selected exemplars are both semantically meaningful and spatially representative, thereby enhancing the reliability of the feedback mechanism.





\begin{figure*}[t]
    \centering
    \includegraphics[width=1\linewidth]{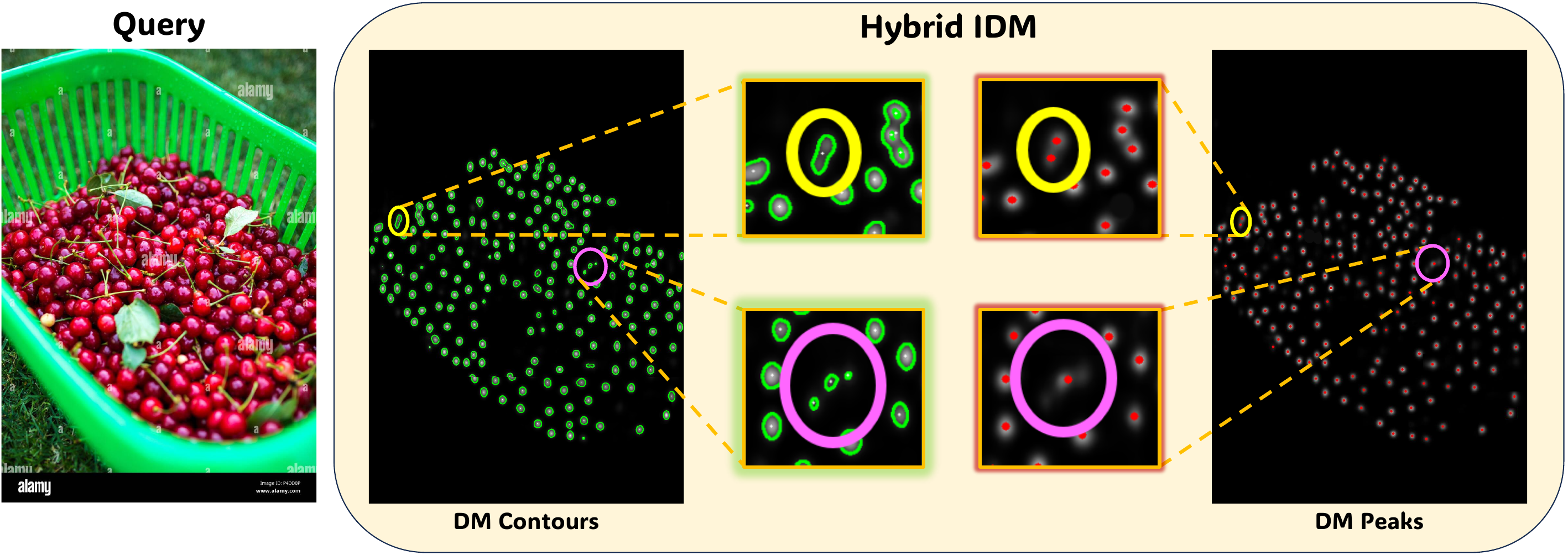}
    \caption{Illustration of the complementary strengths of contour-based and peak-based strategies in Hybrid IDM. The region highlighted with a yellow circle demonstrates the case where contour-based detection fails to resolve multiple closely packed instances, but peak detection accurately identifies distinct object centers. Conversely, the pink circle indicates the area where peak detection fails due to the absence of prominent intensity maxima, yet contour-based separation successfully detects the merged object regions.}

    \label{fig:idmimprove}
\end{figure*}



\noindent\textbf{Scale Diversity: }Each exemplar in the filtered set \( \mathcal{E}_{\text{filtered}} \) has its weighted score \( \Phi \). By sorting the exemplars based on this score, the top \( N \) exemplars can be selected for DM generation. However, the weighted score alone does not guarantee scale diversity in the final selection, which may result in the selection of good exemplars but with similar scales. To ensure scale diversity, we divide the exemplar boxes in the filtered set \( \mathcal{E}_{\text{filtered}} \) into three size categories: small, medium, and large, based on their areas. The categorization assigns each exemplar to a category using a percentile-based criterion that accounts for the distribution of exemplar areas. We compute  33\textsuperscript{rd} and 66\textsuperscript{th} percentiles of the distribution and assign each bounding box \( B_i \) to one of three size categories based on its area \( A_i \). Using percentiles ensures that the categorization is adaptive to the area distribution within each image, making it robust to variations in object sizes across different images.
\[
\text{Category}(A_i) =
\begin{cases}
\text{small}, & \text{if } A_i \leq P_{33}, \\
\text{medium}, & \text{if } P_{33} < A_i \leq P_{66}, \\
\text{large}, & \text{if } A_i > P_{66}.
\end{cases}
\]

\noindent After categorization, the bounding boxes within each size category are sorted in descending order based on their weighted scores \(\Phi(B_i)\). This sorting prioritizes the bounding box with the highest score in each category, capturing both its feature similarity and geometric alignment with the reference exemplar. The top-scoring bounding box from each size category is then selected, ensuring that every size category contributes one exemplar to the final selection.
\begin{equation}
B_{\text{top}}^{\text{C}} = \arg\max_{B_i \in \mathcal{B}_{\text{C}}} \Phi(B_i)
\end{equation}

\noindent where \( \mathcal{B}_{\text{C}} \) denotes the set of bounding boxes in category \( \text{C} \), and \( B_{\text{top}}^{\text{C}} \) is the exemplar with the highest weighted score in category \( \text{C} \). The final selection, \( \mathcal{E}_{\text{final}} = \{B_{\text{top}}^{\text{small}}, B_{\text{top}}^{\text{medium}}, B_{\text{top}}^{\text{large}}\} \) includes exemplars that account for both the feature diversity of the object of interest in the query image and the scale diversity, ensuring a representative set of exemplars.


\subsection{Hybrid IDM}
\label{hybridIDM}

While the IDM in PerSense (Sec.~\ref{subsection:IDM}) relied solely on contour-based centroid detection for generating point prompts from DMs, this approach can sometimes fall short in detecting child instances within complex contiguous density patterns. The core limitation stems from the use of distance transform, which, despite being effective in many scenarios, is still fundamentally a threshold-based method. As a result, it may struggle to accurately separate tightly packed objects in some highly dense or structurally ambiguous regions, where multiple instances form a single tightly merged contour, lacking sufficient boundary distinction.

To address this, PerSense\texttt{++} introduces a hybrid instance detection mechanism that augments the original IDM with a peak detection strategy, thereby enhancing robustness and recall. The newly integrated component identifies local intensity maxima directly from the normalized grayscale DM, enabling the detection of individual object centers even within unbroken high-density blobs that lack distinct contour boundaries. Empirical results show that the two strategies are complementary: contour-based detection is effective in isolating merged regions without prominent peaks, while peak detection excels at resolving closely packed objects within a single contiguous contour (Fig.~\ref{fig:idmimprove}). The final set of candidate point prompts is obtained by taking the union of centroids from both strategies.


\noindent\textbf{Peak Detection via Adaptive Thresholding: }Let \( G(x, y) \in [0, 255] \) denote the normalized grayscale DM derived from the predicted DM \( D(x, y) \). To robustly identify peaks, we compute the mean \( \mu_G \) and standard deviation \( \sigma_G \) of the grayscale map:

\begin{align}
\mu_G &= \frac{1}{|\Omega|} \sum_{(x, y) \in \Omega} G(x, y), \\
\sigma_G &= \sqrt{ \frac{1}{|\Omega|} \sum_{(x, y) \in \Omega} \left(G(x, y) - \mu_G\right)^2 }
\end{align}

\noindent We then define an \textit{adaptive threshold} as:
\begin{equation}
T_{\text{peak}} = \mu_G + \alpha \cdot \sigma_G
\end{equation}
where \( \alpha \in \mathbb{R} \) controls the selectivity of peak detection. A pixel \( (x, y) \) is considered a \textit{local peak} if it satisfies:

\begin{align}
G(x, y) &> T_{\text{peak}} \quad \text{and} \quad G(x, y) > G(x', y')
\end{align}

\noindent \(\forall (x', y') \in \mathcal{N}(x, y) \), where \( \mathcal{N}(x, y) \) denotes the neighborhood around \( (x, y) \). This operation identifies local maxima with intensity significantly above the global distribution, corresponding to potential object centers. Let the resulting set of coordinates from this method be:
\begin{equation*}
\mathcal{Q}_{\text{peak}} = \left\{ (x_i, y_i) \;\middle|\; \text{peak conditions satisfied} \right\}
\end{equation*}


\noindent Finally, the complete set of point prompts in hybrid IDM is constructed by taking the \textit{union} of the original contour-based detections \( \mathcal{Q}_{\text{contour}} \) and the newly introduced peak-based detections:
\begin{equation}
\mathcal{Q}_{\text{hybrid}} = \mathcal{Q}_{\text{contour}} \cup \mathcal{Q}_{\text{peak}}
\end{equation}

\noindent This union ensures that both spatial spread (via contour splitting) and high-density central cues (via peak detection) are captured, thereby reducing missed detections in complex scenes.

\begin{figure}[t]
    \centering
    \includegraphics[width=1\linewidth]{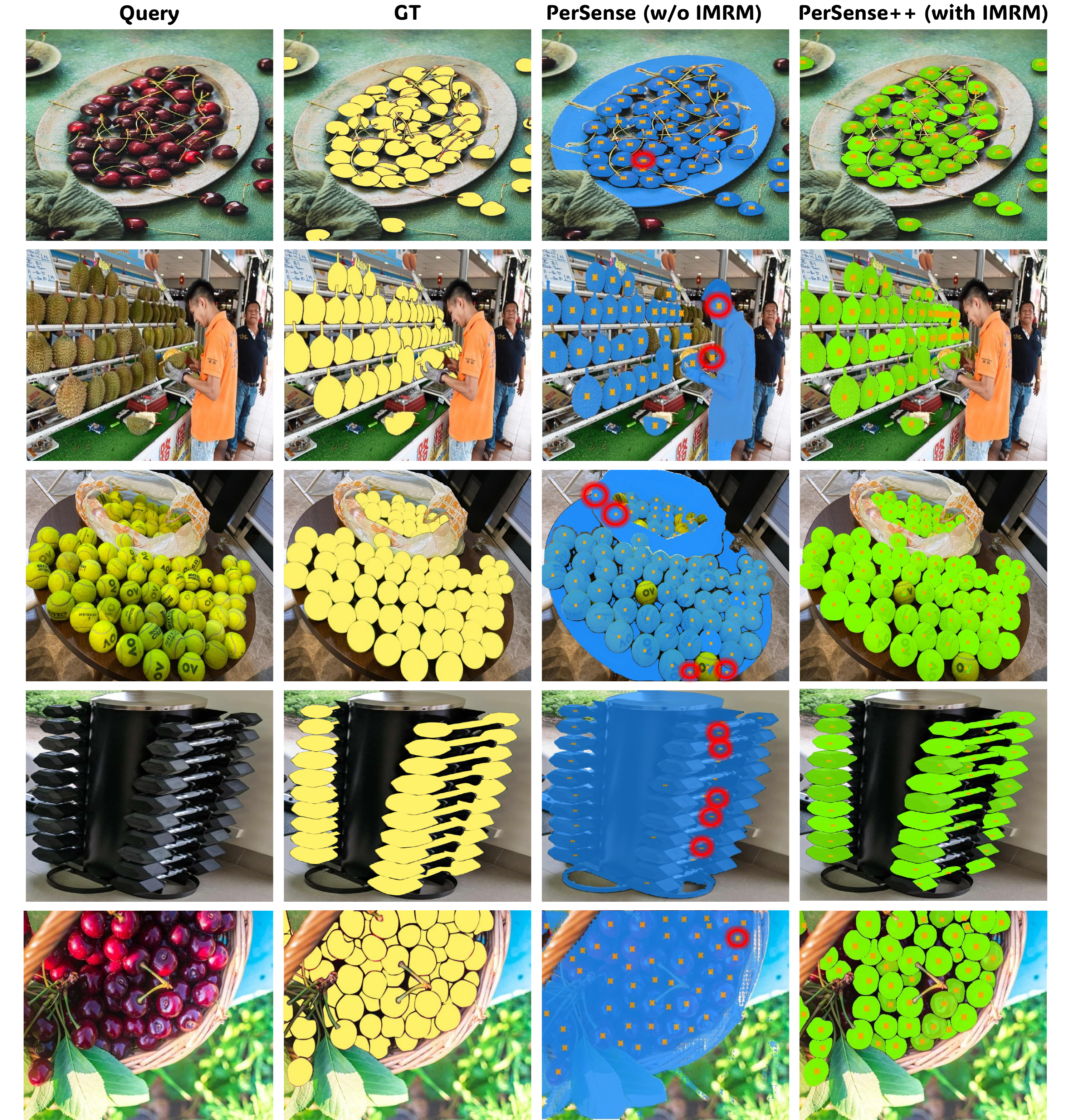}
    \caption{Each row shows an example image and corresponding segmentation outputs. The red circles in the third column highlight spurious prompts triggering irrelevant segmentation in PerSense. IMRM in PerSense\texttt{++} effectively removes these large outlier masks, retaining only consistent, target-class segmentations, thereby enhancing the final segmentation quality.}
    \label{fig:imrm}
\end{figure}

\subsection{Irrelevant Mask Rejection Module (IMRM)}
\label{subsection:imrm}

While PerSense incorporates PPSM to suppress spurious prompts through adaptive thresholding and spatial box-gating, a few false positives may still pass through to the decoder. These residual prompts can generate high-quality but semantically irrelevant masks, such as background regions or objects outside the target class. Left unfiltered, these masks degrade overall segmentation quality by introducing noise into the final prediction.

To address this, we introduce the Irrelevant Mask Rejection Module (IMRM) in PerSense\texttt{++}, a lightweight, post-processing module designed to remove masks that significantly deviate in size from valid object masks. Specifically, IMRM leverages the \textit{area distribution} of the generated masks to identify and reject spatial outliers, particularly very large masks that typically correspond to background or unrelated objects (as illustrated in Fig.~\ref{fig:imrm}). The filtering process in IMRM involves the following steps:

\begin{enumerate}
    \item \textbf{Mask Area Computation:}  
Given a set of binary masks \( \{m_i(x, y)\}_{i=1}^{N} \), each mask area is given by \( a_i = \sum_{x,y} m_i(x, y) \), with \( A = \{a_i\}_{i=1}^{N} \).

    \item \textbf{IQR-Based Cutoff:}  
The 25\textsuperscript{th} and 75\textsuperscript{th} percentiles of \( A \) are used to compute the interquartile range (IQR), i.e., \( Q_1 = \text{percentile}_{25}(A), \quad Q_3 = \text{percentile}_{75}(A) \).
\begin{equation*}
\text{IQR} = Q_3 - Q_1, \quad
T_{\text{IQR}} = Q_3 + 2 \cdot \text{IQR}
\end{equation*}

 \noindent The IQR cutoff offers a robust, non-parametric criterion for filtering extreme values that lie far above the upper quartile, independent of distributional assumptions. It is particularly helpful in handling skewed or heavy-tailed area distributions.

    \item \textbf{Majority Cluster Statistics:}  
    We apply $K$-means clustering to partition the mask areas into two clusters. The cluster with the most members is treated as the \textit{majority cluster}, assumed to represent valid object masks. For this cluster, we compute the mean and standard deviation as:
    \vspace{-3pt}
    \begin{align}
    \mu_{\text{maj}} &= \frac{1}{|M|} \sum_{i \in M} a_i, \\
    \sigma_{\text{maj}} &= \sqrt{\frac{1}{|M|} \sum_{i \in M} (a_i - \mu_{\text{maj}})^2}
    \end{align}

    \noindent The term \( \mu_{\text{maj}} + 2\sigma_{\text{maj}} \) provides a parametric outlier rejection threshold based on the assumption that mask areas in the majority cluster are approximately normally distributed. Under this assumption, values beyond two standard deviations above the mean are considered potential outliers.

    \item \textbf{Combined Threshold and Filtering:}  
    To ensure robustness, we combine both the parametric and non-parametric thresholds. The final area threshold \( T \) is defined as:
    \begin{equation}
    T = (\mu_{\text{maj}} + 2 \cdot \sigma_{\text{maj}}) + T_{\text{IQR}}
    \end{equation}

\end{enumerate}

\begin{figure}[t]
    \centering
    \includegraphics[width=1\linewidth]{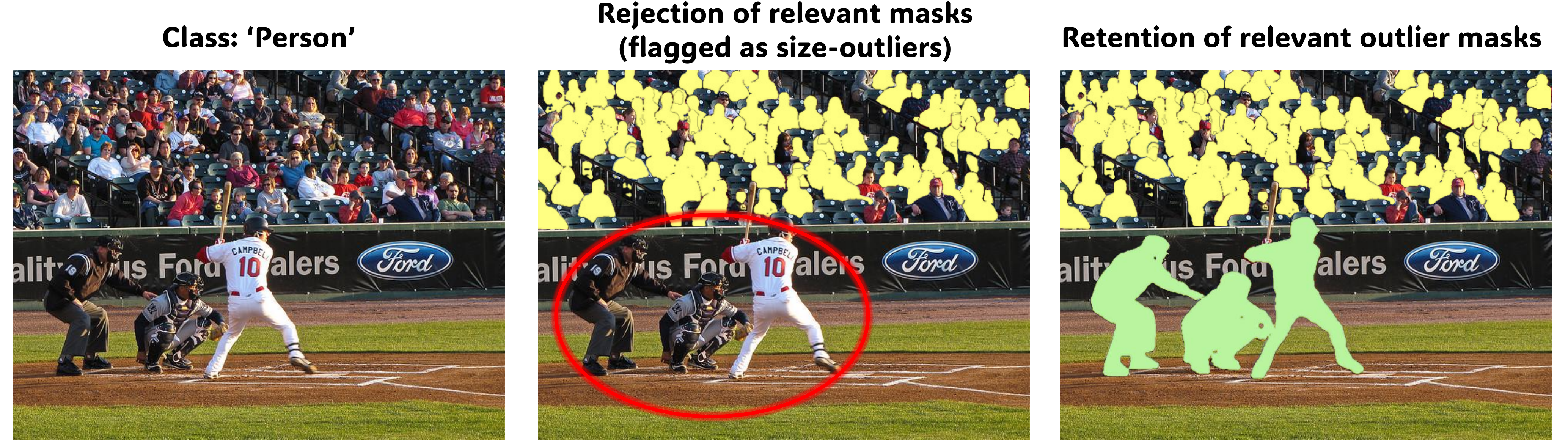}
\caption{The figure illustrates perspective distortion, where valid masks (e.g., \textit{person} instances in red circle) appear larger than the majority cluster (shown in yellow) due to proximity and risk rejection by size-based thresholding. IMRM addresses this by leveraging grounded detections to retain such valid outliers (shown in green).}

    \label{fig:imrm_upd}
\end{figure}

\begin{figure*}[t]
    \centering
    \includegraphics[width=1\linewidth]{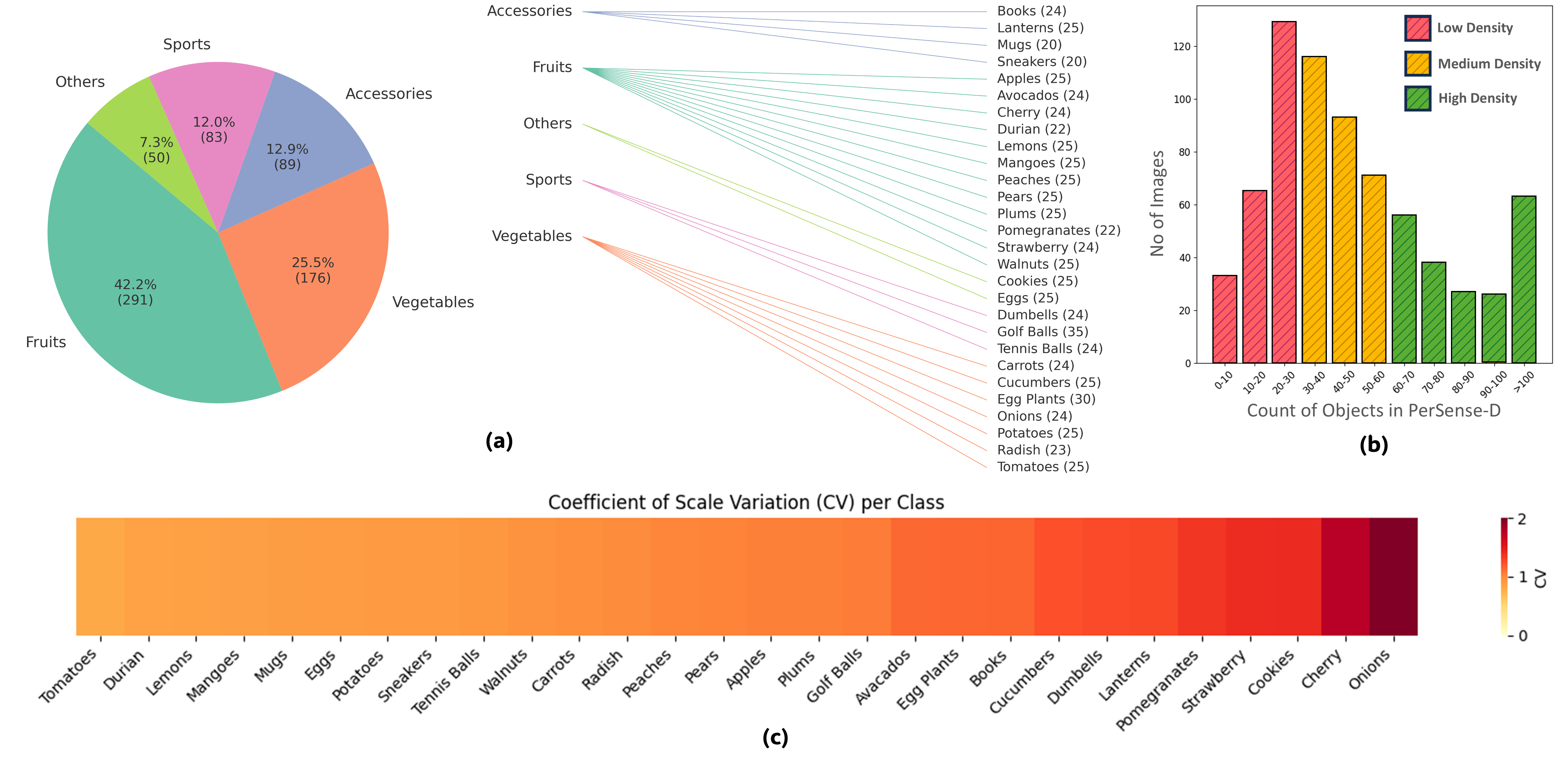}
    \caption{(a) Object categories in PerSense-D. (b) Image distribution across object count bins grouped by density levels. (c) Coefficient of Variation (CV) in object scale across classes, defined as CV = $\sigma / \mu$. Higher values indicate greater intra-class scale variability.}
    \label{fig:dtasetstats}
\end{figure*}

\noindent A mask \( m_i \) is retained if and only if \( a_i \leq T \). This design enables IMRM to identify and reject large, anomalous masks that typically correspond to false-positive point prompts segmenting the background or unrelated objects, examples of which are highlighted with red circles in Fig.~\ref{fig:imrm}. It is important to note that IMRM does not aim to remove masks based on semantic context, as false-positive prompts may produce masks that partially overlap with target objects. In such cases, feature-based filtering using reference embeddings may fail to reject these ambiguous masks due to shared visual features. Additionally, if we compute the similarity between the reference and candidate masks using cosine similarity, a common choice for comparing high-dimensional features, it primarily measures directional alignment in feature space rather than true semantic correspondence. As a result, high similarity scores can be obtained for masks covering background or irrelevant regions, especially when those regions partially include the target object. Applying a threshold on such noisy similarity scores can be detrimental, potentially leading to the rejection of valid masks or the retention of irrelevant ones.

To address this, IMRM focuses purely on spatial outlier detection based on area distribution. However, this introduces a limitation under perspective distortion, where objects closer to the camera naturally appear larger than those farther away. In such scenarios, if the majority of target instances are small being farther in the scene, a valid object appearing large (due to proximity) may be incorrectly flagged as an outlier and rejected. To mitigate this, we leverage bounding box detections from grounding detector. For each mask flagged as a size-based outlier, we compute its IoU with the grounded detection boxes. If the flagged outlier exhibits a sufficiently high overlap (e.g., IoU $\geq$ 0.8) with any detection, it is retained. This safeguards IMRM from mistakenly discarding large but valid masks due to perspective effects. See Fig.~\ref{fig:imrm_upd} for illustration.



\section{New Evaluation Benchmark (PerSense-D)}
\label{sec:benchmark}

Existing segmentation datasets like COCO~\cite{lin2014microsoft}, LVIS~\cite{gupta2019lvis}, and FSS-1000~\cite{li2020fss} include multi-instance images but rarely represent dense scenes due to limited object counts. For example, LVIS averages only 3.3 instances per category. To bridge this gap, we introduce PerSense-D, \textit{a dense segmentation benchmark of 717 images spanning across 28 object categories}   (Fig.~\ref{fig:dtasetstats}). On average, each image contains 53 object instances, with a total of 36,837 annotated objects across the entire dataset. The number of instances per image ranges from 7 to 573, and the average image resolution is 839 \(\times\) 967 pixels.In addition to dense layouts, PerSense-D exhibits high intra-class object scale diversity, as reflected by the coefficient of variation (CV) in object scale across categories (Fig.~\ref{fig:dtasetstats}). The CV is computed as the ratio of the standard deviation to the mean of normalized instance areas (scale) within each category, providing a size-invariant measure of relative dispersion. This enhances the dataset’s value as a challenging testbed for advancing segmentation methods in real-world dense scenarios.

\noindent\textbf{Image Collection and Retrieval:} To collect the dense set, we searched for object categories across Google, Bing, and Yahoo, using prefixed keywords like "\textit{multiple}", "\textit{lots of}", and "\textit{many}" to encourage dense results. For each of the 28 categories, we retrieved the top 100 images, yielding 2800 candidates for further filtering.

\noindent\textbf{Manual Inspection and Filtering:}  
Candidate images were manually filtered based on the following criteria: (1) Adequate quality for clear object distinction, (2) at least 7 object instances per image, following dense counting datasets~\cite{ranjan2021learning}, and (3) presence of dense, cluttered scenes with occlusions. This resulted in 689 images from an initial pool of 2800.

\noindent\textbf{Semi-automatic Image Annotation Pipeline:}  
We crowdsourced annotations using a semi-automatic pipeline. Following a model-in-the-loop approach~\cite{kirillov2023segment}, PerSense generated initial masks, which annotators refined using OpenCV and Photoshop tools for pixel-accurate corrections. Given the high instance count,  manual refinement averaged 15 minutes per image. We also provide dot annotations at object centers.

\noindent\textbf{Evaluation Protocol:} To ensure fairness under one-shot setting, we provide a set of 28 support images (labeled “00”), each containing a single object instance per category. This eliminates randomness in support selection and standardizes evaluation across methods. PerSense-D supports both class-wise and density-based evaluations for granular performance analysis. For the latter, images are categorized into three density levels based on object count: \textit{Low} (\(C_I \leq 30\), 224 images), \textit{Medium} (\(30 < C_I \leq 60\), 266 images), and \textit{High} (\(C_I > 60\), 199 images). Fig.~\ref{fig:dtasetstats} shows image distribution by object count and density category.

\section{Experiments}\label{experiments}
\vspace{-1pt}
\subsection{Implementation Details}
\label{subsection:implementation_details}

Our PerSense and PerSense\texttt{++} are model-agnostic and leverage a CLE, grounding detector, and DMG for personalized instance segmentation in dense images. For CLE, we leverage VIP-LLaVA~\citep{cai2024vipllava}, which utilizes CLIP-336px~\citep{radford2021learning} and Vicuna v1.5~\citep{chiang2023vicuna} as visual and language encoders, respectively. For grounding, we employ GroundingDINO~\citep{liu2023grounding} as the detection module, adjusting the detection threshold to 15\%. 
To demonstrate model-agnostic capability of PerSense, we separately utilize DSALVANet~\citep{he2024few} and CounTR~\citep{liu2022countr} pretrained on FSC-147~\citep{ranjan2021learning} as DMG1 and DMG2, respectively. Finally, to ensure fair comparison with other approaches, we utilize vanilla SAM~\citep{kirillov2023segment} encoder and decoder. We use standard evaluation metric of mean Intersection over Union (mIoU) for evaluating segmentation performance. To assess the accuracy of the generated DMs, particularly when comparing PerSense with PerSense\texttt{++}, we report Mean Absolute Error (MAE) and Root Mean Square Error (RMSE), following established object counting protocols~\citep{huang2024point, pelhan2024dave}. \textbf{\textit{No training is involved in any of our experiments.}}

\subsection{Datasets}
\label{subsection:datasets}

We evaluate both class-wise and density-based segmentation performance on the proposed PerSense-D benchmark, adhering to the evaluation protocol described in Sec.~\ref{sec:benchmark}. Additionally, we report results on standard segmentation benchmarks, including COCO~\citep{lin2014microsoft} and LVIS~\citep{gupta2019lvis}. For COCO-20$^{i}$~\citep{nguyen2019feature}, the 80 object categories in COCO are divided into four cross-validation folds, each comprising 60 training classes and 20 test classes. Unlike specialist models, PerSense is \textit{NOT} trained on this dataset; instead, we assess its performance solely on the designated test classes. For LVIS-92$^{i}$, we follow the evaluation scheme introduced in~\cite{liu2023matcher}.

Beyond overall evaluation on the entire COCO-20\textsuperscript{i} test set, which consists of approximately 41k images, we further analyze performance on its subset with dense images only. Following the image selection criterion in PerSense-D, we identify images containing at least seven instances of a specific class. This filtering yields a new subset COCO-20\textsuperscript{d}, consisting of 9184 dense images from the COCO-20\textsuperscript{i} test set. For LVIS-92\textsuperscript{i}, it results in a new subset LVIS-92\textsuperscript{d}, comprising of 7204 dense images from the original 20k test set.


\begin{table*}[t] 
\centering

\setlength{\tabcolsep}{4.8pt} 
\fontsize{7pt}{7pt}\selectfont
\begin{tabular}{llc|c|cccc} 
\toprule
 &  & \multicolumn{3}{c}{\textbf{Density-based}} & \textbf{Class-wise} \\ 
\textbf{Method} & \textbf{Venue} & \multicolumn{3}{c}{\textbf{(mIoU)}} & \textbf{(mIoU)} \\ 
\cmidrule(lr){3-5} \cmidrule(lr){6-6}
& & \textbf{Low} & \textbf{Med} & \textbf{High} & \textbf{Overall} \\ 
\midrule

\multicolumn{6}{l}{\underline{\textit{End-to-end trained / fine-tuned}}} \\
\noalign{\vskip 2pt}
C3Det~\citep{lee2022interactive} & CVPR 2022 & 52.70 & 46.64 & 39.11 & 48.60 \\
\noalign{\vskip 2pt}
SegGPT~\citep{wang2023seggpt} & ICCV 2023 & 59.81 & 53.34 & 52.05 & 55.50 \\
PerSAM-F$^\dagger$~\citep{zhang2023personalize} & ICLR 2024 & 38.18 & 34.84 & 26.73 & 29.30 \\
PseCo$^\dagger$~\citep{huang2024point} & CVPR 2024 & 53.99 & 65.23 & 68.55 & 61.83 \\
GeCo$^\dagger$~\citep{pelhannovel} & NIPS 2024 & 63.92 & 63.40 & 74.49 & 65.95 \\

\midrule
\multicolumn{6}{l}{\underline{\textit{Training-free}}} \\
\noalign{\vskip 2pt}
PerSAM$^\dagger$~\citep{zhang2023personalize} & ICLR 2024 & 32.27 & 28.75 & 20.25 & 24.45 \\
TFOC$^\dagger$~\citep{shi2024training} & WACV 2024 & \underline{62.78} & 65.38 & 65.69 & 62.63 \\
Matcher$^\dagger$~\citep{liu2023matcher} & ICLR 2024 & 58.62 & 58.30 & 68.00 & 62.80 \\
GroundedSAM$^\dagger$~\citep{ren2024grounded} & arXiv 2024 & 58.36 & 66.24 & 64.97 & 65.92 \\
\textbf{PerSense}$^\dagger$ (DMG1) & (ours) & \textbf{66.36} & \underline{67.27} & \underline{74.78} & \underline{70.96} \\
\textbf{PerSense}$^\dagger$ (DMG2) & (ours) & 59.84 & \textbf{73.51} & \textbf{77.57} & \textbf{71.61} \\

\bottomrule
\end{tabular}
\caption{Comparison of PerSense with SOTA segmentation approaches on PerSense-D benchmark. We report density-based evaluation along with class-wise performance following the evaluation protocol in Sec.~\ref{sec:benchmark}.  $^\dagger$ indicates method using SAM.}
\label{tab:persense_comparison}
\end{table*}

\begin{table*}[t] 
\centering
\setlength{\tabcolsep}{4.3pt} 
\fontsize{7pt}{7pt}\selectfont
\begin{tabular}{llc|c|c|c|cc} 
\toprule
\multirow{3}{*}{\textbf{Method}} & \multirow{3}{*}{\textbf{Venue}} & \multicolumn{5}{c}{\textbf{COCO-20\textsuperscript{i}}} & \textbf{LVIS-92\textsuperscript{i}} \\ 
\cmidrule(lr){3-7} \cmidrule(lr){8-8}
& & \textbf{F0} & \textbf{F1} & \textbf{F2} & \textbf{F3} & \textbf{\shortstack{Mean \\ mIoU}} & \textbf{\shortstack{Mean \\ mIoU}} \\ 
\midrule

\multicolumn{8}{l}{\underline{\textit{In-domain training}}} \\ 
\noalign{\vskip 2pt}

HSNet~\citep{min2021hypercorrelation} & CVPR 21 & 37.2 & 44.1 & 42.4 & 41.3 & 41.2 & 17.4 \\
VAT~\citep{hong2022cost} & ECCV 22 & 39.0 & 43.8 & 42.6 & 39.7 & 41.3 & 18.5 \\
FPTrans~\citep{zhang2022feature} & NIPS 22 & 44.4 & 48.9 & 50.6 & 44.0 & 47.0 & - \\
MIANet~\citep{yang2023mianet} & CVPR 23 & 42.4 & 52.9 & 47.7 & 47.4 & 47.6 & - \\
LLaFS~\citep{zhu2024llafs} & CVPR 24 & 47.5 & 58.8 & 56.2 & 53.0 & 53.9 & - \\

\midrule
\multicolumn{8}{l}{\underline{\textit{COCO as training data}}} \\
\noalign{\vskip 2pt}
Painter~\citep{wang2023images} & CVPR 23 & 31.2 & 35.3 & 33.5 & 32.4 & 33.1 & 10.5 \\
SegGPT~\citep{wang2023seggpt} & ICCV 23 & 56.3 & 57.4 & 58.9 & 51.7 & 56.1 & 18.6 \\

\midrule
\multicolumn{8}{l}{\underline{\textit{Training-free (excluding PerSAM-F~\citep{zhang2023personalize})}}} \\
\noalign{\vskip 2pt}
PerSAM$^\dagger$~\citep{zhang2023personalize} &ICLR 24  & 23.1 & 23.6 & 22.0 & 23.4 & 23.0 & 11.5 \\
PerSAM-F$^\dagger$~\citep{zhang2023personalize} & ICLR 24 & 22.3 & 24.0 & 23.4 & 24.1 & 23.5 & 12.3 \\
Matcher$^\dagger$~\citep{liu2023matcher} &ICLR 24  & \textbf{52.7} & \textbf{53.5} & \textbf{52.6} & \textbf{52.1} & \textbf{52.7} & \textbf{33.0} \\
\textbf{PerSense}$^\dagger$ & (ours) &  \underline{47.8} &  \underline{49.3} &  \underline{48.9} &  \underline{50.1} &  \underline{49.0} &  \underline{25.7} \\


\bottomrule
\end{tabular}
\caption{Comparison of PerSense with SOTA approaches on COCO-20\textsuperscript{i} and LVIS-92\textsuperscript{i}. $^\dagger$ indicates method using SAM.}
\label{tab:comparison}
\end{table*}

\begin{table*}[t]
\centering
\setlength{\tabcolsep}{4.8pt}
\fontsize{7pt}{7pt}\selectfont

\begin{tabular}{ll|c|c|c|c|c|c}
\toprule
\multirow{3}{*}{\textbf{Method}} & \multirow{3}{*}{\textbf{Venue}} & \multicolumn{5}{c|}{\textbf{COCO-20\textsuperscript{d}}} & \textbf{LVIS-92\textsuperscript{d}} \\
\cmidrule(lr){3-7}
\cmidrule(lr){8-8}
 & & F0 & F1 & F2 & F3 & Mean mIoU & mIoU \\
 & & (5359 imgs) & (1369 imgs) & (1511 imgs) & (945 imgs) & (9184 imgs) & (7204 imgs) \\

\multicolumn{8}{l}{\underline{\textit{Training-free}}} \\
PerSAM~\citep{zhang2023personalize} & ICLR 2024 & 13.3 & 14.1 & 17.4 & 16.5 & 15.3 & 13.5 \\
Matcher~\citep{liu2023matcher} & ICLR 2024 & 31.1 & 28.0 & 35.5 & 33.0 & 31.9 & 18.4 \\
\textbf{PerSense} & (Ours) & \textbf{56.1} & \textbf{54.7} & \textbf{62.6} & \textbf{61.5} & \textbf{58.7} & \textbf{28.7} \\

\bottomrule
\end{tabular}

\caption{mIoU comparison of PerSense with SOTA on COCO-20\textsuperscript{d} and LVIS-92\textsuperscript{d}. The number of images in each fold is indicated in parentheses.}
\label{tab:persense_dense}
\end{table*}


\subsection{Results and Discussion}
\label{subsection:results}

\subsubsection{PerSense vs SOTA}
\label{subsection:persenseresults}

\textbf{Performance on PerSense-D:} Table~\ref{tab:persense_comparison} presents performance of PerSense in dense scenarios as compared to SOTA approaches utilizing PerSense-D as an evaluation benchmark. PerSense achieves \textbf{71.61\%} overall class-wise mIoU, outperforming one-shot in-context segmentation approaches, including PerSAM~\citep{zhang2023personalize} (\textbf{\texttt{+}47.16\%}), PerSAM-F~\citep{zhang2023personalize} (\textbf{\texttt{+}42.27\%}), SegGPT~\citep{wang2023seggpt} (\textbf{\texttt{+}16.11\%}) and Matcher~\citep{liu2023matcher} (\textbf{\texttt{+}8.83\%}). PerSAM's decline stems from premature segmentation termination due to its naive confidence thresholding, which misidentifies remaining instances as background in dense scenes. For Matcher, we observed that in dense scenarios, the patch-level matching and correspondence matrix struggles to identify distinct regions when there is significant overlap or occlusion among objects. To be fair in comparison with Grounded-SAM, we ensured that all classes in PerSense-D overlap with the datasets used for GroundingDINO pretraining (notably, all classes are present in Objects365 dataset~\citep{shao2019objects365}). 
PerSense surpassed Grounded-SAM by \textbf{\texttt{+}5.69\%}, demonstrating its robustness.

Since object counting methods primarily target dense scenes, we compare PerSense with the recently introduced training-free object counting framework (TFOC)~\citep{shi2024training}, which formulates the counting task as a segmentation problem. PerSense outperforms TFOC by \textbf{\texttt{+}8.98\%}. Additionally, we evaluate PerSense against SOTA counting approaches PseCo~\citep{huang2024point} and GeCo~\citep{pelhannovel} which essentially are dense object detectors end-to-end trained on FSC-147 dataset. PerSense outperforms PseCo by \textbf{\texttt{+}9.78\%} and GeCo by \textbf{\texttt{+}5.66\%}, demonstrating its superiority. Finally, we compare PerSense with C3Det~\citep{lee2022interactive}, a multi-class tiny object detector trained on Tiny-DOTA dataset~\cite{xia2018dota}. Using the positive location prior from PerSense as the initial user input, C3Det was tasked to detect similar instances, which were then segmented using SAM. PerSense outperformed C3Det by \textbf{\texttt{+}23.01\%}.



For a fine-grained analysis, we additionally evaluate PerSense's segmentation performance across varying object densities, following the PerSense-D density-based categorization discussed in Sec.~\ref{sec:benchmark}. We compute mIoU separately for low, medium, and high-density images, where PerSense outperforms all methods, establishing itself as the SOTA for personalized segmentation in dense images. An interesting observation is the monotonic increase in PerSense's segmentation performance from low to high density images. This is due to DMs generated by DMG, which are inherently suited for structured object distributions in dense environments. In low-density images, object sparsity limits high-confidence regions, making point prompt generation challenging. As density increases, the DM becomes richer, enabling IDM to extract more precise points, thereby improving segmentation accuracy in denser scenarios.

\begin{table*}[t]
\centering
\setlength{\tabcolsep}{5pt}
\fontsize{7pt}{7pt}\selectfont

\begin{tabular}{ll|ccc|c|c|c}
\toprule
\textbf{Method} & \textbf{Venue} 
& \multicolumn{3}{c|}{\textbf{Density-based (mIoU)}} 
& \multicolumn{1}{c|}{\textbf{Class-wise (mIoU)}} 
& \textbf{MAE $\downarrow$} & \textbf{RMSE $\downarrow$} \\

\cmidrule(lr){3-5} \cmidrule(lr){6-6}
& & \textbf{Low} & \textbf{Med} & \textbf{High} & \textbf{Overall} & & \\ 
\midrule

\textbf{PerSense} (DMG1) & (Ours) & 66.36 & 67.27 & 74.78 & 70.96 & 15.07 & 30.35 \\
\textbf{PerSense\texttt{++}} (DMG1) & (Ours) & 70.87 & 75.86 & 81.93 & \underline{77.45} & \underline{12.91} & \underline{26.17} \\
\textbf{PerSense} (DMG2) & (Ours) & 59.84 & 73.51 & 77.57 & 71.61 & 16.17 & 40.76 \\
\textbf{PerSense\texttt{++}} (DMG2) & (Ours) & 74.82 & 82.00 & 83.94 & \textbf{81.35} & \textbf{11.48} & \textbf{24.67} \\

\bottomrule
\end{tabular}

\caption{Comparison of PerSense and PerSense\texttt{++} on the PerSense-D benchmark. We report density-based and class-wise mIoU, along with DM accuracy using MAE and RMSE metrics.}
\label{tab:persensevspersense++Persense-D}
\end{table*}

\begin{table*}[t]
\centering
\setlength{\tabcolsep}{4.3pt}
\fontsize{7pt}{7pt}\selectfont

\begin{tabular}{ll|cccc|c|cccc|c|c|c}
\toprule
\textbf{Method} & \textbf{Venue} 
& \multicolumn{5}{c|}{\textbf{COCO-20\textsuperscript{i}}} 
& \multicolumn{5}{c|}{\textbf{COCO-20\textsuperscript{d}}} 
& \textbf{LVIS-92\textsuperscript{i}} 
& \textbf{LVIS-92\textsuperscript{d}} \\

\cmidrule(lr){3-7} \cmidrule(lr){8-12} \cmidrule(lr){13-13} \cmidrule(lr){14-14}
& & F0 & F1 & F2 & F3 & \textbf{Mean} 
  & F0 & F1 & F2 & F3 & \textbf{Mean} 
  & \textbf{mIoU} & \textbf{mIoU} \\
\midrule

\textbf{PerSense}$^\dagger$ & (Ours) 
& 47.8 & 49.3 & 48.9 & 50.1 & 49.0 
& 56.1 & 54.7 & 62.6 & 61.5 & 58.7 
& 25.7 & 28.7 \\

\textbf{PerSense\texttt{++}}$^\dagger$ & (Ours) 
& \textbf{48.4} & \textbf{50.2} & \textbf{49.5} & \textbf{51.0} & \textbf{49.7} 
& \textbf{58.5} & \textbf{56.9} & \textbf{64.7} & \textbf{63.2} & \textbf{60.8} 
& \textbf{26.9} & \textbf{31.0} \\

\bottomrule
\end{tabular}

\caption{Comparison of PerSense and PerSense++ on COCO-20\textsuperscript{i/d} and LVIS-92\textsuperscript{i/d}.}
\label{tab:persense++vspersensecocolvis}
\end{table*}

\begin{table*}[t]
\centering
\setlength{\tabcolsep}{4pt}
\fontsize{7pt}{7pt}\selectfont

\begin{tabular}{l|c|c|c|c|c|c|c|c|c}
\toprule
\textbf{Method} 
& \textbf{Grounded-SAM} 
& \textbf{PerSAM} 
& \textbf{Matcher} 
& \multicolumn{3}{c|}{\textbf{PerSense}} 
& \multicolumn{3}{c}{\textbf{PerSense\texttt{++}}} \\
& & & 
& \textbf{N = 1} & \textbf{N = 16} & \textbf{N = 32}
& \textbf{N = 1} & \textbf{N = 16} & \textbf{N = 32} \\
\midrule
\textbf{Memory (MB)} 
& 2943 & 2950 & 3209 
& 2988 & 3060 & 3125 
& 3037 & 3109 & 3174 \\
\textbf{Avg Inf Time (sec)} 
& 1.8 & \(C \times 1.02\) & 10.2 
& 2.7 & 1.6 & 1.5  
& 3.0 & 1.9 & 1.8 \\
\bottomrule
\end{tabular}

\caption{Runtime and memory usage comparison. PerSense and PerSense\texttt{++} are evaluated at varying point prompt batch sizes \(N \in \{1, 16, 32\}\). For PerSAM, \(C\) indicates the number of instances in the image.}
\label{tab:runtime_efficiency}
\end{table*}




\begin{figure}[t]
    \centering
    \includegraphics[width=1\linewidth]{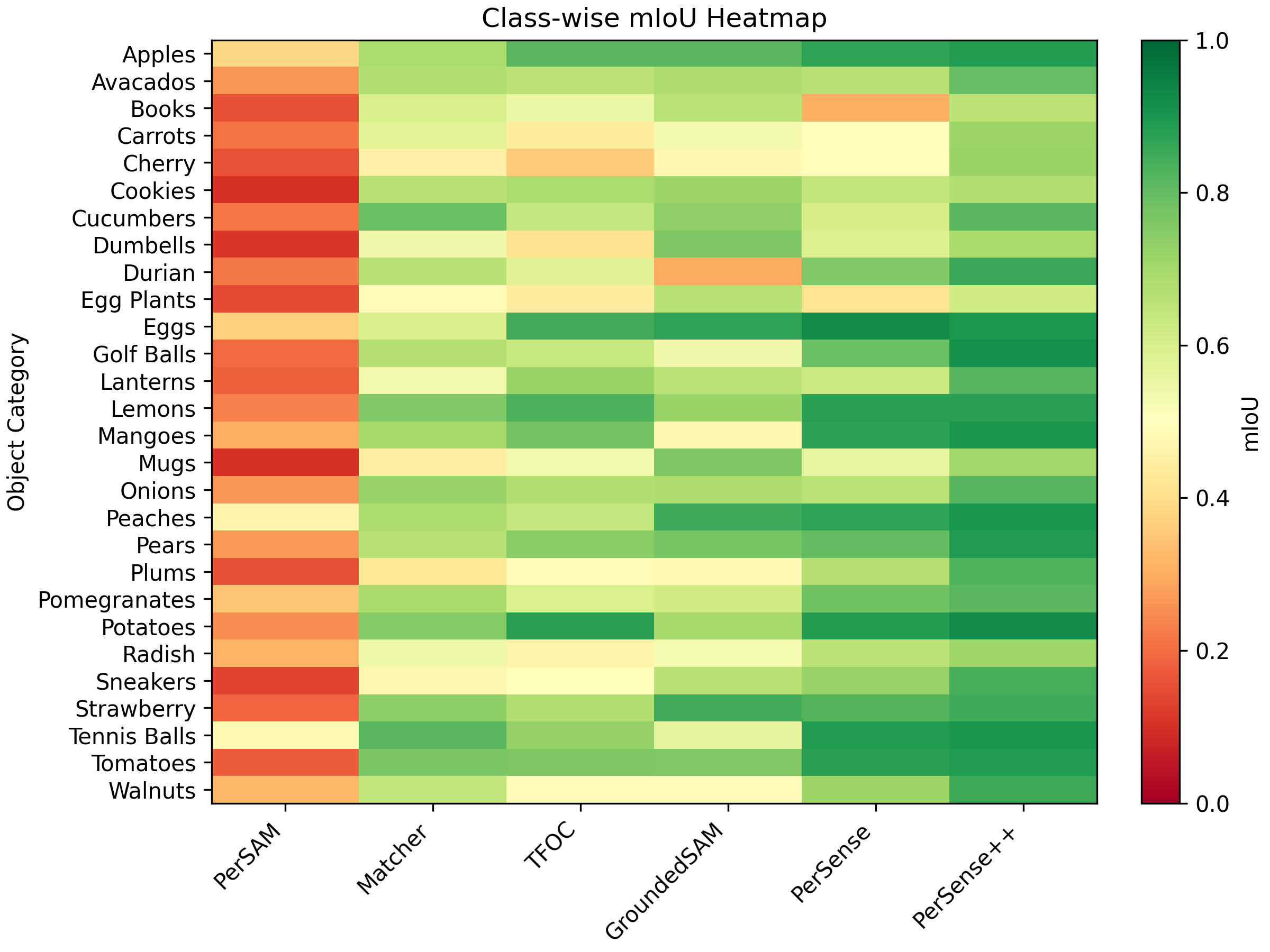}
\caption{Class-wise mIoU comparison of training-free methods on the PerSense-D benchmark. The heatmap clearly highlights the improvement in segmentation accuracy achieved by PerSense\texttt{++} over PerSense and other SOTA methods.}

    \label{fig:classwiseheatmap} 
\end{figure}

\begin{figure}[t]
    \centering
    \includegraphics[width=1\linewidth]{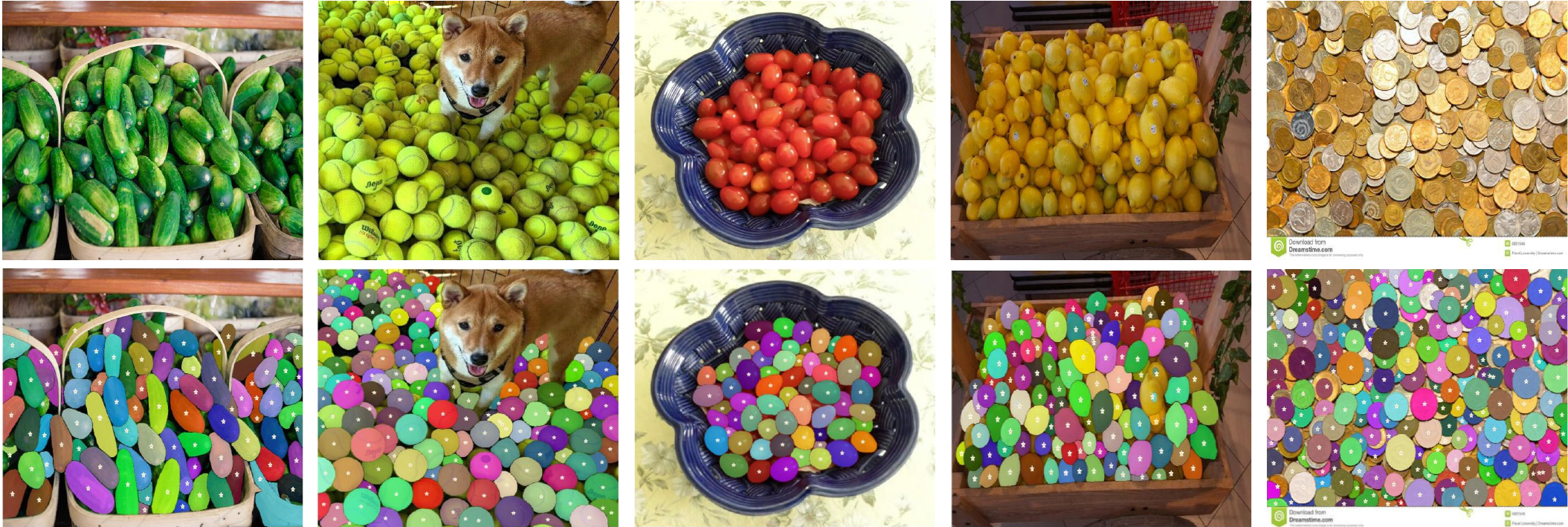}
    \caption{PerSense \textit{in-the-wild}. $1^{\text{st}}$ row: \textit{Query Image}, $2^{\text{nd}}$ row: \textit{Instance Segmentation}.}
    \label{fig:inthewild}
\end{figure}

\begin{figure*}[t]
    \centering
    \includegraphics[width=1\linewidth]{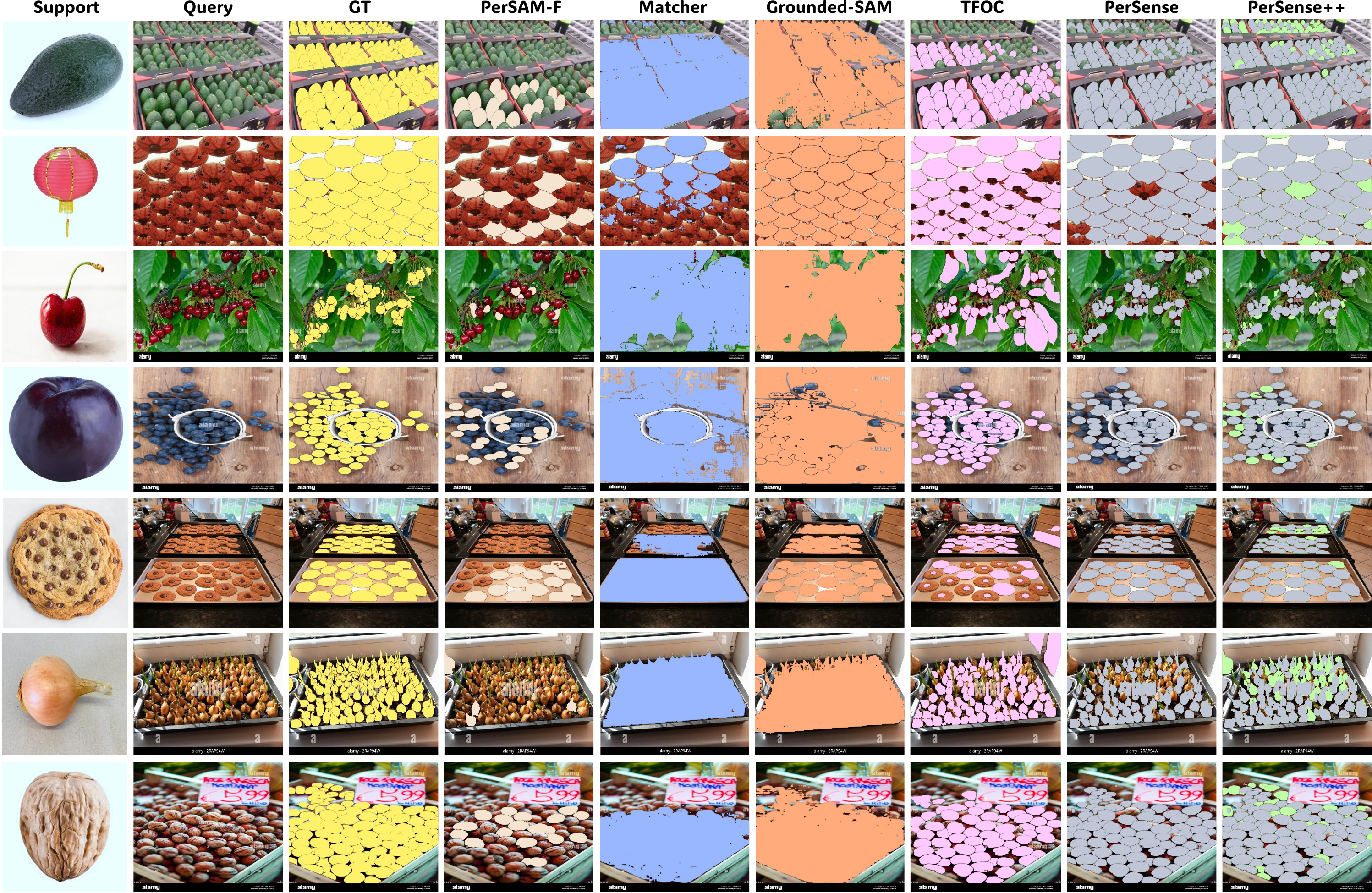}
    \caption{Qualitative comparison of PerSense\texttt{++} with PerSense and other SOTA approaches on PerSense-D benchmark. The improvement offered by PerSense\texttt{++} over PerSense is highlighted in green.}
    \label{fig:qualitative_results_persense++_persense-D} 
\end{figure*}

\noindent\textbf{Generalization on Standard Benchmarks:} PerSense is specifically designed for dense images, deriving point prompts from DMs generated by DMG. As a result, it is naturally not expected to give SOTA performance on sparse images with few object instances, e.g. LVIS averages 3.3 instances per category. For images with single or low object count, DM tends to spread across the entire object, reducing its effectiveness and undermining its primary purpose of capturing variations in object density across the image. In such cases, traditional object detection methods are more effective due to fewer occlusions and easier object boundary delineation, rendering DM generation inefficient. Nonetheless, to provide a comprehensive evaluation, we evaluate PerSense on COCO-20\textsuperscript{i}~\citep{nguyen2019feature} and LVIS-92\textsuperscript{i}~\citep{gupta2019lvis}, demonstrating that despite being tailored for dense scenes, PerSense maintains competitive performance even in sparse settings. To provide a broader perspective, we compare PerSense with both in-domain training methods and training-free approaches (Table~\ref{tab:comparison}). Despite being a training-free framework, PerSense achieves performance comparable to several well-known in-domain training methods. 
PerSense demonstrates significant improvements over PerSAM-F, achieving mIoU gains of \textbf{\texttt{+}25.5\%} on COCO-20\textsuperscript{i} and \textbf{\texttt{+}13.4\%} on LVIS-92\textsuperscript{i}.  PerSense outperforms SegGPT~\citep{wang2023seggpt} on LVIS-92\textsuperscript{i} by \textbf{\texttt{+}7.1\%}. However, SegGPT demonstrates superior performance on COCO-20\textsuperscript{i}, as it is included in its training set. Additionally, PerSense surpasses Painter~\citep{wang2023images} on COCO-20\textsuperscript{i} and LVIS-92\textsuperscript{i} by \textbf{\texttt{+}15.9\%} and \textbf{\texttt{+}15.2\%} mIoU, respectively, despite Painter being trained on COCO. Overall, PerSense ranks second-best among SOTA, trailing only Matcher~\citep{liu2023matcher}.  This highlights PerSense robustness and generalization ability, even in scenarios outside its primary design focus.

To highlight the effectiveness of PerSense in dense scenes corresponding to existing segmentation benchmarks, we report results on COCO-20\textsuperscript{d} and LVIS-92\textsuperscript{d}: dense subsets of COCO-20\textsuperscript{i} and LVIS-92\textsuperscript{i} test sets, as described in Sec.~\ref{subsection:datasets}. As shown in Table~\ref{tab:persense_dense}, PerSense delivers substantial performance improvements, surpassing PerSAM and Matcher by \textbf{\texttt{+}43.42\%} and \textbf{\texttt{+}26.84\%} on COCO-20\textsuperscript{d}, respectively. On LVIS-92\textsuperscript{d}, it achieves mIoU gains of \textbf{\texttt{+}15.19\%} and \textbf{\texttt{+}10.28\%} over PerSAM and Matcher, respectively. These consistent improvements across both proposed and existing benchmarks position PerSense as SOTA in one-shot segmentation for dense images. We further evaluate \textit{in-the-wild} performance of PerSense by scraping random images from internet as illustrated in Fig.~\ref{fig:inthewild}.

\begin{figure*}[t]
    \centering
    \includegraphics[width=1\linewidth]{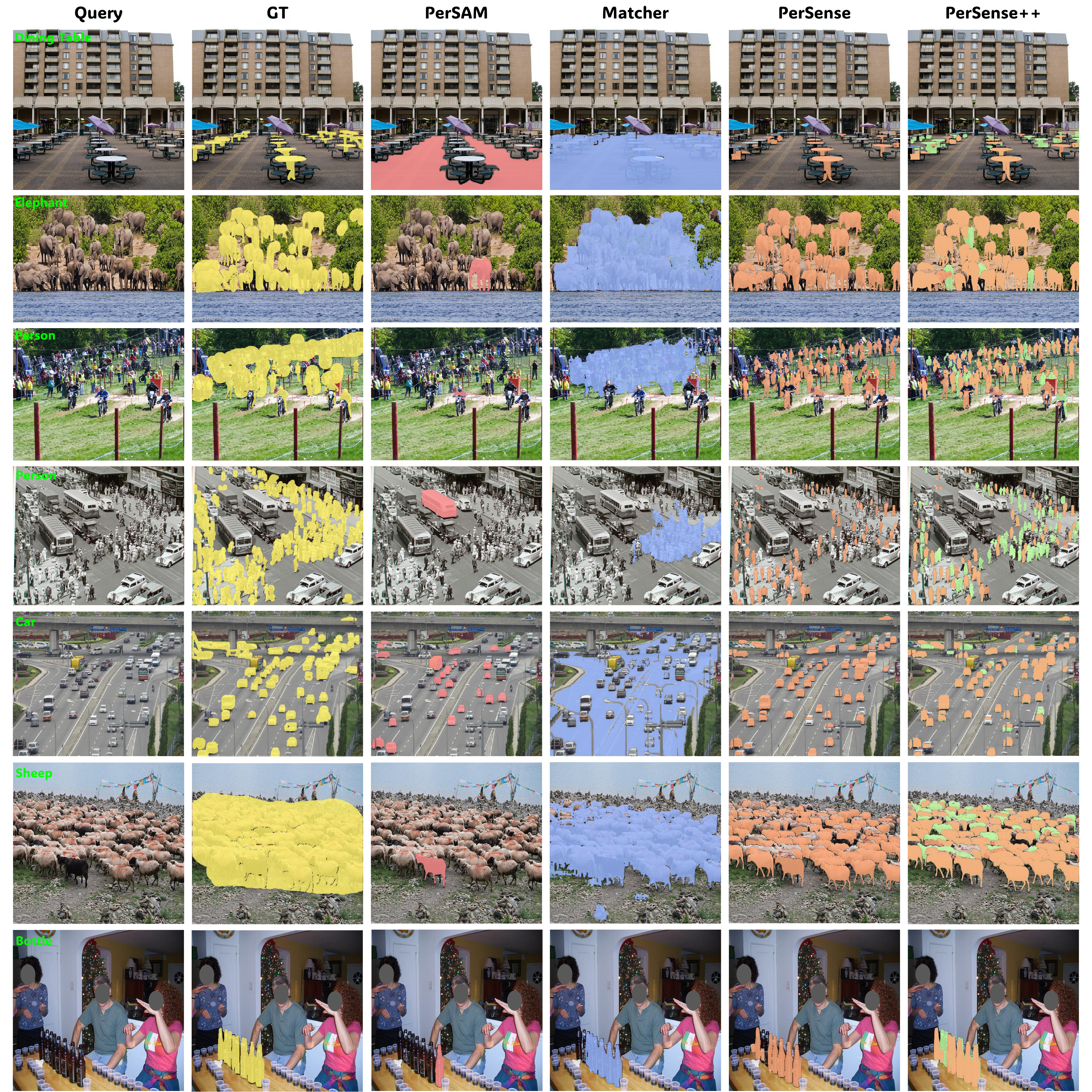}
    \caption{Qualitative comparison of PerSense\texttt{++} with PerSense and other SOTA approaches on  COCO-20\textsuperscript{d} and LVIS-92\textsuperscript{d}. The improvement offered by PerSense\texttt{++} over PerSense is highlighted in green.}
    \label{fig:qualitative_results_persense++_cocolvis} 
\end{figure*}

\subsubsection{PerSense\texttt{++} vs PerSense}
\label{subsection:persense++results}

PerSense\texttt{++}, which integrates a diversity-aware exemplar selection strategy, Hybrid IDM, and IMRM, consistently outperforms PerSense across all evaluated datasets. On the PerSense-D benchmark, it achieves a substantial class-wise mIoU gain of \textbf{\texttt{+}6.49\%} for DMG1 and \textbf{\texttt{+}9.7\%} for DMG2 (Table~\ref{tab:persensevspersense++Persense-D}). Class-wise comparison among training-free methods on the PerSense-D benchmark is presented in Fig.~\ref{fig:classwiseheatmap}, while qualitative comparisons with PerSense and other SOTA methods are shown in Fig.~\ref{fig:qualitative_results_persense++_persense-D}.

On COCO-20\textsuperscript{i} and LVIS-92\textsuperscript{i}, PerSense\texttt{++} improves over PerSense by \textbf{\texttt{+}0.7\%} and \textbf{\texttt{+}1.2\%} mIoU, respectively (Table~\ref{tab:persense++vspersensecocolvis}). In addition, it achieves gains of \textbf{\texttt{+}2.1\%} and \textbf{\texttt{+}2.3\%} on the dense subsets COCO-20\textsuperscript{d} and LVIS-92\textsuperscript{d}, respectively (Table~\ref{tab:persense++vspersensecocolvis}). These improvements on dense subsets exceed those observed on the full test sets, emphasizing the strength of PerSense\texttt{++} in handling crowded and cluttered scenes. Qualitative results on dense subsets are presented in Fig.~\ref{fig:qualitative_results_persense++_cocolvis}. It is important to note that the ground truth masks provided for dense images in COCO and LVIS are not instance-level; instead, they are coarse, blob-like masks representing dense regions. This underrepresents the performance of methods like PerSense and PerSense\texttt{++}, which predict fine-grained instance-level masks. Despite this sub-optimal ground truth, our approach still significantly outperforms existing SOTA methods.

\begin{figure}[h]
    \centering
    \includegraphics[width=1\linewidth]{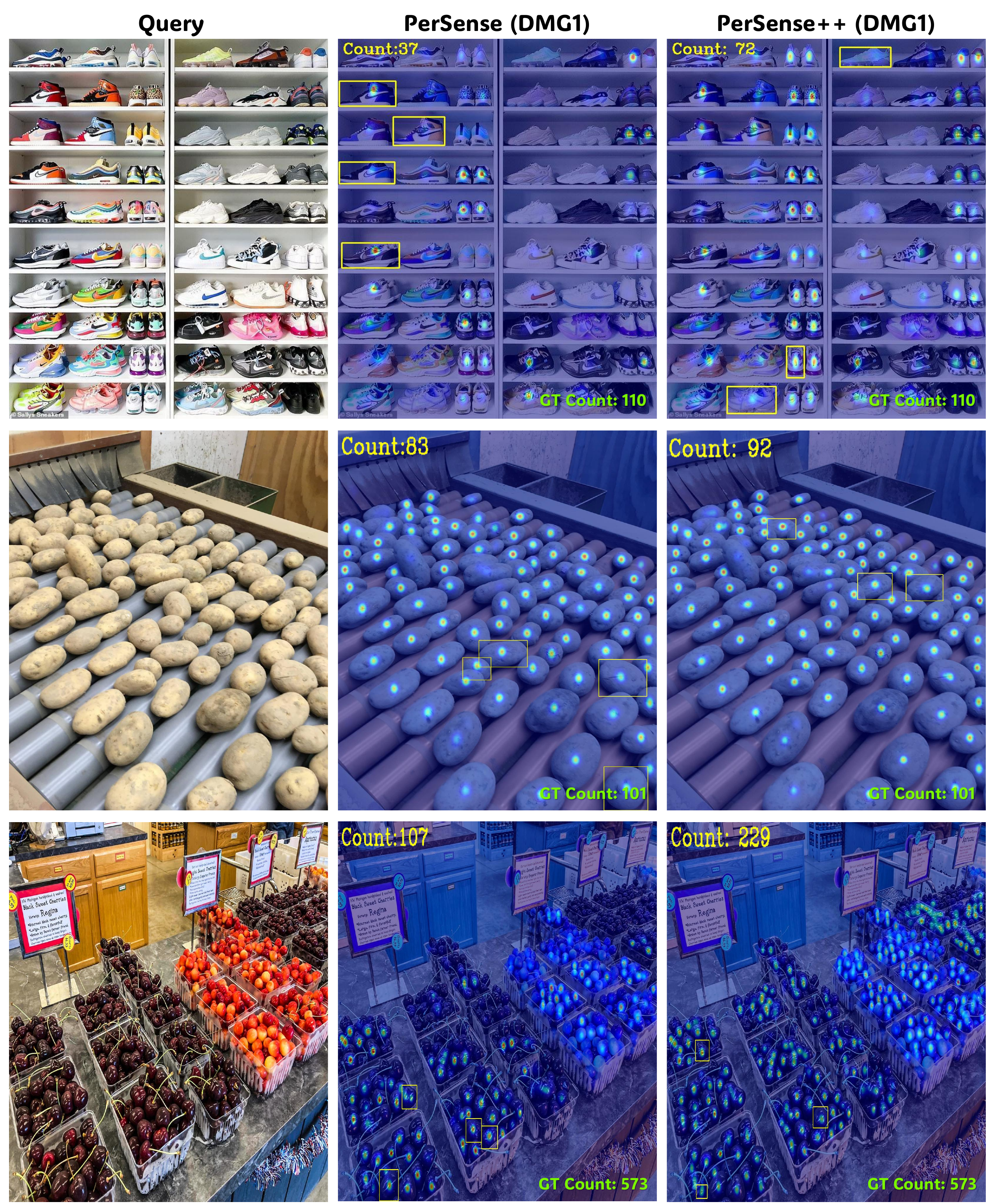}
\caption{Qualitative analysis of DM accuracy improvements on PerSense-D. PerSense\texttt{++} demonstrates superior counting performance compared to PerSense, despite using only three exemplars versus four in the original PerSense.}

    \label{fig:countingperformance} 
\end{figure}

Importantly, PerSense\texttt{++} achieves these gains using only three exemplars compared to four exemplars in PerSense, highlighting the efficiency of its improved exemplar selection strategy. Since segmentation performance in our framework is closely tied to DM accuracy, we further assess the impact of the diversity-aware exemplar selection strategy on object counting. As shown in Table~\ref{tab:persensevspersense++Persense-D}, PerSense\texttt{++} enhances DM accuracy on the PerSense-D benchmark, which directly contributes to improved segmentation performance in dense scenes. Qualitative improvement in DM accuracy is illustrated in Fig.~\ref{fig:countingperformance}.


\begin{figure}[t]
    \centering
    \includegraphics[width=1\linewidth]{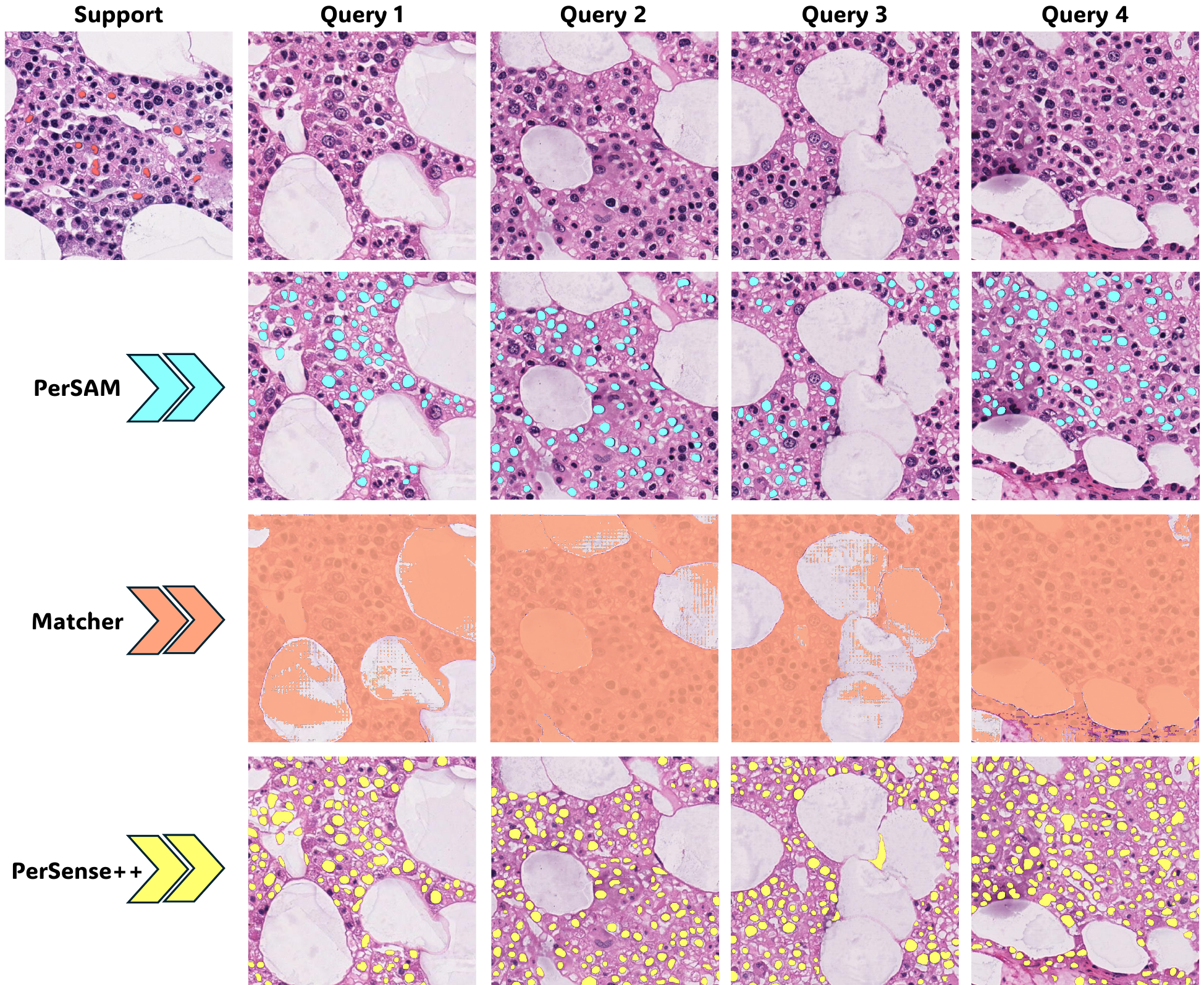}
\caption{Model-agnostic capability of PerSense\texttt{++} demonstrated on the MBM bone marrow cell counting dataset. For fairness, the vanilla SAM is replaced with the MedSAM encoder and decoder in PerSAM, Matcher, and PerSense\texttt{++}. PerSense\texttt{++} consistently outperforms SOTA one-shot segmentation methods, showcasing adaptability to specialized domains.}
    \label{fig:medical} 
\end{figure}


\noindent\textbf{Runtime Efficiency: } Table~\ref{tab:runtime_efficiency} reports the inference time and memory consumption of PerSense and PerSense\texttt{++}, evaluated on a single NVIDIA GeForce RTX 4090 GPU with an image batch size of 1. To further enhance runtime efficiency, we process the point prompts generated for each image in batches. In other words, segmentation masks for $N$ point prompts are generated in parallel, as each prompt is processed independently. Our framework demonstrates superior runtime efficiency compared to PerSAM and Matcher and is comparable to Grounded-SAM in terms of inference speed. It incurs slightly higher GPU memory usage than Grounded-SAM, though with a significant trade-off in segmentation accuracy. Furthermore, as a model-agnostic framework, it offers the flexibility to integrate lightweight backbones tailored to the target application, enabling users to balance between accuracy and efficiency based on deployment requirements.

\noindent\textbf{Model-Agnostic Framework: }In addition to demonstrating model-agnostic capability with two different DMGs, PerSense also generalizes to dense scenarios in the medical domain. Fig.~\ref{fig:medical} presents qualitative results for cellular segmentation  on Modified Bone Marrow (MBM) dataset~\citep{paul2017count} leveraging MedSAM~\citep{ma2024segment} encoder and decoder instead of vanilla SAM. These results highlight PerSense's broad applicability, spanning domains from industrial to medical.

\begin{table*}[t]
  \raggedright
  \setlength{\tabcolsep}{3pt}
  \fontsize{6pt}{6pt}\selectfont

  \begin{minipage}[t]{0.44\textwidth} 
    \centering
    \subcaption{}
    \label{tab:component-wise}
    \setlength{\tabcolsep}{1.5pt} 
    \begin{tabular}{lccc}
      \toprule
      \multirow{2}{*}{\textbf{Modules}} & \textbf{baseline} & \textbf{baseline \texttt{+} PPSM} & \textbf{PerSense} \\
      & \textbf{(mIoU)} & \textbf{(mIoU)} & \textbf{(mIoU)} \\
      \midrule
      IDM & yes & yes & yes \\
      PPSM & no & yes & yes \\ 
      Feedback & no & no & yes \\
      \midrule
      PerSense-D \textcolor{blue}{(Gain)} & 65.58 \textcolor{blue}{(-)} & 66.95 \textcolor{blue}{(+1.37)} & 70.96 \textcolor{blue}{(+4.01)} \\
      COCO \textcolor{blue}{(Gain)} & 46.33 \textcolor{blue}{(-)} & 48.81 \textcolor{blue}{(+2.48)} & 49.00 \textcolor{blue}{(+0.19)} \\
      \bottomrule
    \end{tabular}
  \end{minipage}%
  \hspace{0.015\textwidth}
  \begin{minipage}[t]{0.11\textwidth} 
    \centering
    \subcaption{}
    \label{tab:norm-fact}
    \setlength{\tabcolsep}{1.5pt} 
    \begin{tabular}{cc}
      \toprule
      \textbf{Norm} & \textbf{mIoU} \\
      \textbf{Factor} & \\
      \midrule
      1 & 70.41 \\
      $\sqrt{2}$ & 70.96 \\
      $\sqrt{3}$ & 69.59 \\
      $\sqrt{5}$ & 68.95 \\
      \bottomrule
    \end{tabular}
  \end{minipage}%
  \hspace{0.015\textwidth}
  \begin{minipage}[t]{0.11\textwidth} 
    \centering
    \captionsetup[sub]{labelformat=empty}
    \subcaption{\hspace{10pt}(c)}
    \label{dmg}
    \setlength{\tabcolsep}{1.7pt} 
    \begin{tabular}{cc}
      \toprule
      \textbf{No. of} & \textbf{mIoU} \\
      \textbf{Exemplars} & \\
      \midrule
      1 & 65.78 \\
      2 & 69.24 \\
      3 & 70.53 \\
      4 & 70.96 \\
      5 & 70.90 \\
      6 & 70.81 \\
      \bottomrule
    \end{tabular}
  \end{minipage}%
  \hspace{0.035\textwidth}
  \begin{minipage}[t]{0.12\textwidth} 
    \centering
    \captionsetup[sub]{labelformat=empty}
    \subcaption{\hspace{45pt}(d)}
    \label{iterations}
    \setlength{\tabcolsep}{1.2pt}
    \begin{tabular}{ccccc}
      \toprule
      \textbf{No. of} & 1 & 2 & 3 & 4 \\
      \textbf{Iterations} & & & & \\
      \midrule
      \textbf{PerSense} & \multirow{2}{*}{70.96} & \multirow{2}{*}{70.97} & \multirow{2}{*}{70.96} & \multirow{2}{*}{70.95} \\
      \textbf{(mIoU)} & & & & \\
      \midrule
      \textbf{Avg Inf} & \multirow{2}{*}{2.7} & \multirow{2}{*}{3.1} & \multirow{2}{*}{3.5} & \multirow{2}{*}{3.9} \\
      \textbf{time (sec)} & & & & \\
      \bottomrule
    \end{tabular}
  \end{minipage}%

  \caption{(a) Component-wise ablation study of PerSense. (b) Choice of normalization factor for adaptive threshold in PPSM. (c) Varying number of exemplars in DMG using feedback mechanism. (d) Impact of multiple feedback iterations on PerSense performance.}
  \label{merged-ablation-tables}
\end{table*}

\subsection{Ablation Study}\label{ablations}

\subsubsection{PerSense Ablations}\label{persense_ablations}

\textbf{Component-wise Ablation: } 
The PerSense framework comprises three key components: IDM, PPSM, and a feedback mechanism. An ablation study presented in Table~\ref{tab:component-wise} quantifies their respective contributions. Integrating PPSM into our baseline network improved mIoU by \textbf{+1.37\%} on PerSense-D and \textbf{+2.48\%} on COCO, effectively mitigating false positives in IDM-generated point prompts. The feedback mechanism further increased mIoU by \textbf{+4.01\%} on PerSense-D, demonstrating its effectiveness in optimizing exemplar selection. However, its impact on COCO was limited (\textbf{+0.19\%}) due to insufficient object instances, restricting its ability to refine the DM.

\noindent\textbf{Choice of Normalization Factor in PPSM: } For the PPSM adaptive threshold, we initialized the normalization constant at 1 and applied a square root progression for a gradual step-size increase. Empirical results show that $\sqrt{2}$ yields the most significant mIoU improvement (Table~\ref{tab:norm-fact}).


\begin{table*}[t]
\centering
\setlength{\tabcolsep}{6pt}
\fontsize{7pt}{7pt}\selectfont

\begin{tabular}{ccccc}
\toprule
\multirow{2}{*}{\textbf{Framework}} & \textbf{Diversity Aware-ES} & \textbf{Hybrid IDM} & \textbf{IMRM} & \textbf{PerSense-D} \\
 & \textbf{(A)} & \textbf{(B)} & \textbf{(C)}& \textbf{(mIoU)}\textcolor{blue}{(Gain)} \\
\midrule
PerSense & No & No & No & 71.61 \\
PerSense \texttt{+} (A)  & Yes & No & No & 75.41 \textcolor{blue}{(+3.80)} \\
PerSense \texttt{+} (A) \texttt{+} (B) & Yes & Yes & No & 78.13 \textcolor{blue}{(+2.72)} \\
PerSense\texttt{++} & Yes & Yes & Yes & \textbf{81.35} \textcolor{blue}{(+3.22)} \\
\bottomrule
\end{tabular}

\caption{Component-wise ablation study of PerSense\texttt{++}}
\label{tab:ablation_persensepp_framework}
\end{table*}

\begin{table}[t]
\centering
\setlength{\tabcolsep}{2.5pt}
\fontsize{7pt}{7pt}\selectfont

\begin{tabular}{ccc|cc}
\toprule
\multicolumn{3}{c|}{\textbf{Diversity-Aware Exemplar Selection}} & \multicolumn{2}{c}{\textbf{DM Accuracy}} \\
\cmidrule(lr){1-3} \cmidrule(lr){4-5}
\textbf{Feature} & \textbf{Geometric} & \textbf{Scale} & \textbf{MAE} $\downarrow$ & \textbf{RMSE} $\downarrow$ \\
\textbf{Diversity} & \textbf{Attributes} & \textbf{Diversity} & & \\
\midrule
Yes & No & No & 13.31 & 31.35 \\
Yes & Yes & No & 12.60 & 29.19 \\
Yes & Yes & Yes & \textbf{11.48} & \textbf{24.67} \\
\bottomrule
\end{tabular}

\caption{Effect of incorporating feature, geometric, and scale-based diversity in exemplar selection on DM accuracy.}
\label{tab:diversity_ablation}
\end{table}

\begin{table*}[t]
\centering
\setlength{\tabcolsep}{6pt}
\fontsize{7pt}{7pt}\selectfont

\begin{tabular}{ccccccc}
\toprule
\textbf{Method} & \textbf{Onions} & \textbf{Cherry} & \textbf{Cookies} & \textbf{Strawberry} & \textbf{Pomegranates} & \textbf{Lanterns} \\
\midrule
PerSAM        & 26.2 & 15.8 & 10.1 & 18.8 & 35.1 & 18.1 \\
Matcher       & 72.1 & 45.6 & 66.4 & 73.5 & 68.9 & 53.3 \\
TFOC          & 67.5 & 36.2 & 68.5 & 67.5 & 59.1 & 71.7 \\
GroundedSAM   & 68.0 & 47.4 & \textbf{71.2} & 84.6 & 61.6 & 65.9 \\
PerSense++    & \textbf{81.8} & \textbf{71.6} & 67.3 & \textbf{84.9} & \textbf{81.6} & \textbf{81.9} \\
\bottomrule
\end{tabular}

\caption{mIoU comparison on PerSense-D classes with high intra-class object scale variation.}
\label{tab:scale_var}
\end{table*}

\noindent\textbf{Varying No. of Exemplars in Feedback Mechanism: } We automated the selection of the best exemplars for DMG based on SAM scores using the proposed feedback mechanism. As shown in Table~\ref{dmg}, segmentation performance on PerSense-D saturates after four exemplars, as additional exemplars do not provide any new significant information about the object of interest. 

\noindent\textbf{Multiple Iterations (Feedback Mechanism):} The PerSense feedback mechanism refines the DM in a single pass by selecting exemplars from the initial segmentation output based on SAM scores. We analyzed multiple iterations (Table~\ref{iterations}) and found no accuracy gains but increased computational overhead. This occurs because first-pass exemplars, with well-defined boundaries, are already effectively captured by SAM. Subsequent iterations redundantly select the same exemplars due to their distinct features and consistently high SAM scores.

\subsubsection{PerSense\texttt{++} Ablations}\label{persense++_ablations}

\noindent\textbf{Component-wise Ablation:} The PerSense\texttt{++} framework enhances the original PerSense by introducing three key components: a diversity-aware exemplar selection strategy, Hybrid IDM, and IMRM. Table~\ref{tab:ablation_persensepp_framework} quantifies the individual contribution of each component to the overall mIoU improvement. Integrating the diversity-aware exemplar selection into PerSense yields a \textbf{+3.80\%} mIoU gain on the PerSense-D benchmark, indirectly reflecting improved DM accuracy through the selection of a more representative and diverse exemplar set. The Hybrid IDM further boosts performance by \textbf{+2.72\%}, leveraging a complementary combination of contour-based and peak-based centroid detection. Finally, IMRM contributes an additional \textbf{+3.22\%} gain by effectively identifying and discarding spatial outlier masks, which typically correspond to background regions or irrelevant objects.

\noindent\textbf{Diversity-aware Exemplar Selection:}
PerSense\texttt{++} integrates a diversity-aware exemplar selection strategy within its feedback mechanism, introducing both feature diversity and scale diversity into the selection process. To assess the individual contributions of these components, we conducted an ablation study focusing on DM accuracy, evaluated through object counting performance using MAE and RMSE metrics. As shown in Table~\ref{tab:diversity_ablation}, incorporating geometric attributes into the feature-based selection, via a weighted scoring mechanism, enhances DM accuracy. This improvement is further amplified when scale diversity is introduced alongside feature diversity, leading to the generation of more precise point prompts, which are then forwarded to the decoder for segmentation.

\noindent\textbf{Performance under High Intra-Class Scale Variation:} Based on the coefficient of object scale variation (CV) reported in Sec.~\ref{sec:benchmark}, we selected the top six classes with the highest CV and present segmentation results in Table~\ref{tab:scale_var}. PerSense++ demonstrates SOTA performance even in dense scenarios with significant object scale variation.

\begin{figure}[t]
    \centering
    \includegraphics[width=1\linewidth]{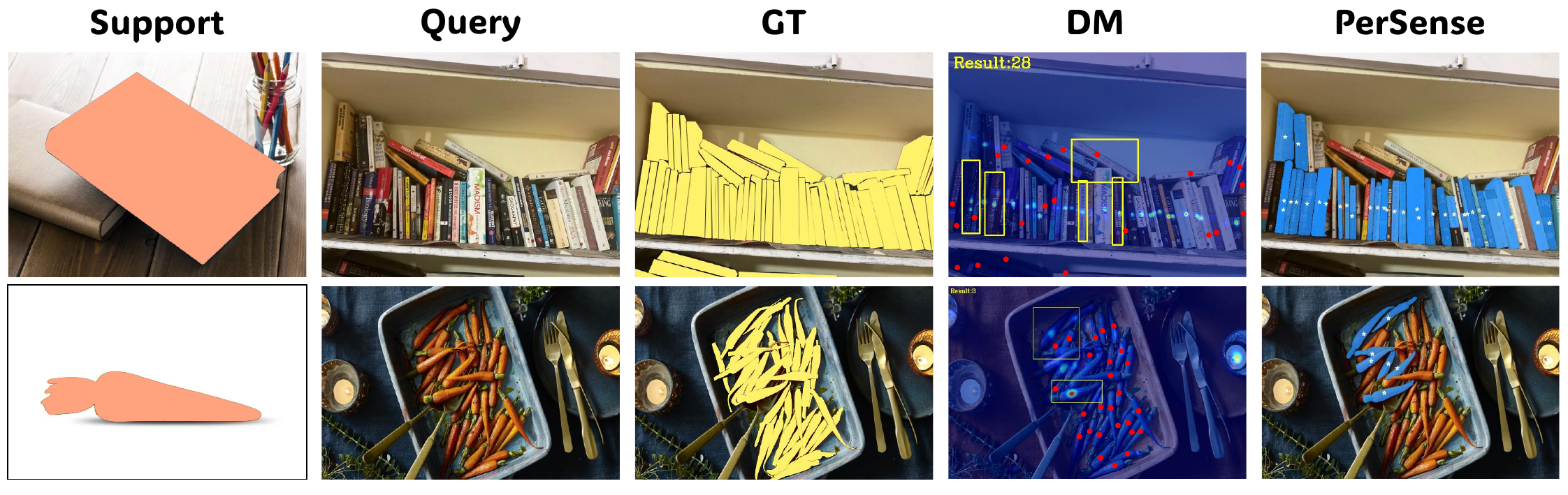}
    \caption[PerSense limitation]{The figure illustrates scenarios where PerSense's performance deteriorates, primarily due to its reliance on the generated DM. In the first row, where the goal is to segment all instances of the "book" class, the DM excludes many true positives (highlighted in red), which PerSense cannot recover once they are lost during DM generation. A similar issue is seen in the second row, where a poor-quality DM for the "carrot" class leads to missed instances, negatively impacting PerSense segmentation performance.}
    \label{fig:failurecases}
\end{figure}

\subsection{Limitations}\label{limitations}

A key limitation of both PerSense and PerSense\texttt{++} lies in their dependence on the quality of the generated DM. Although modules like IDM and PPSM effectively refine the DM and suppress false positives, they cannot recover any true positives that are missed during the initial DM generation by the DMG (Figure~\ref{fig:failurecases}).  While the introduction of the diversity-aware exemplar selection strategy in PerSense\texttt{++} significantly improves DM quality, any true positives that are still not captured in the DM remain unrecoverable. This highlights the critical role of accurate DM generation in ensuring reliable downstream segmentation.


\section{Conclusion}\label{conclusion}
We presented PerSense and its enhanced variant PerSense\texttt{++}: training-free, model-agnostic frameworks for personalized instance segmentation in dense visual scenes. Motivated by real-world applications such as automated quality control in food processing and cellular analysis in medical imaging, our approach addressed key challenges in dense environments, including occlusion, clutter, and scale variation. PerSense leveraged DMs for point prompt generation and refinement, while PerSense\texttt{++} introduced exemplar diversity, hybrid instance detection, and outlier mask rejection to further improve segmentation robustness. Extensive evaluations, supported by the new PerSense-D benchmark, demonstrated that PerSense\texttt{++} outperforms existing methods in both accuracy and efficiency, highlighting its practical value in dense, real-world scenarios.




\backmatter

\bibliography{sn-bibliography}

\end{document}